\documentclass{svproc}
\usepackage{url}

\usepackage{graphicx}
\usepackage{amsmath,epsfig}
\usepackage{booktabs}
\usepackage{amssymb}

\def\et{\textit{et al.}}

\usepackage{bbding}
\usepackage[marginal]{footmisc}

\usepackage{color,xcolor}

\begin{document}
\mainmatter              % start of a contribution
\title{Wider and Higher: Intensive Integration and Global Foreground Perception for Image Matting}
\titlerunning{Intensive Integration and Global Foreground Perception for Image Matting}  % abbreviated title (for running head)
\author{Yu Qiao\inst{1} \and Ziqi Wei\inst{2},
Yuhao Liu\inst{1} \and Yuxin Wang\inst{1(}\Envelope\inst{)} \and Dongsheng Zhou\inst{3} \\ 
Qiang Zhang\inst{1} \and Xin Yang\inst{1}}
\authorrunning{Yu Qiao and Ziqi Wei et al.} % abbreviated author list (for running head)
\tocauthor{Yu Qiao, Ziqi Wei, Yuhao Liu, Yuxin Wang, Dongsheng Zhou, Qiang Zhang, Xin Yang}
\institute{Computer Science and Technology, Dalian University of Technology, Dalian, China,\\
\email{wyx@dlut.edu.cn}
\and
CAS Key Laboratory of Molecular Imaging, Institute of Automation, \\
Beijing, China
\and
Dalian University, Dalian, China}

\maketitle              % typeset the title of the contribution

\footnote{Yu Qiao and Ziqi Wei contribute equally to this work.\\}

\begin{abstract}
This paper reviews recent deep-learning-based matting research and conceives our wider and higher motivation for image matting. Many approaches achieve alpha mattes with complex encoders to extract robust semantics, then resort to the U-net-like decoder to concatenate or fuse encoder features. However, image matting is essentially a pixel-wise regression, and the ideal situation is to perceive the maximum opacity correspondence from the input image. In this paper, we argue that the high-resolution feature representation, perception and communication are more crucial for matting accuracy. Therefore, we propose an Intensive Integration and Global Foreground Perception network (I2GFP) to integrate wider and higher feature streams. Wider means we combine intensive features in each decoder stage, while higher suggests we retain high-resolution intermediate features and perceive large-scale foreground appearance. Our motivation sacrifices model depth for a significant performance promotion. We perform extensive experiments to prove the proposed I2GFP model, and state-of-the-art results can be achieved on different public datasets.
% We would like to encourage you to list your keywords within
% the abstract section using the \keywords{...} command.
\end{abstract}

\section{Introduction}
\label{sec:intro}
\begin{figure}[t]
	\begin{center}
		\begin{tabular}{cc}
			\includegraphics[width=0.46\linewidth]{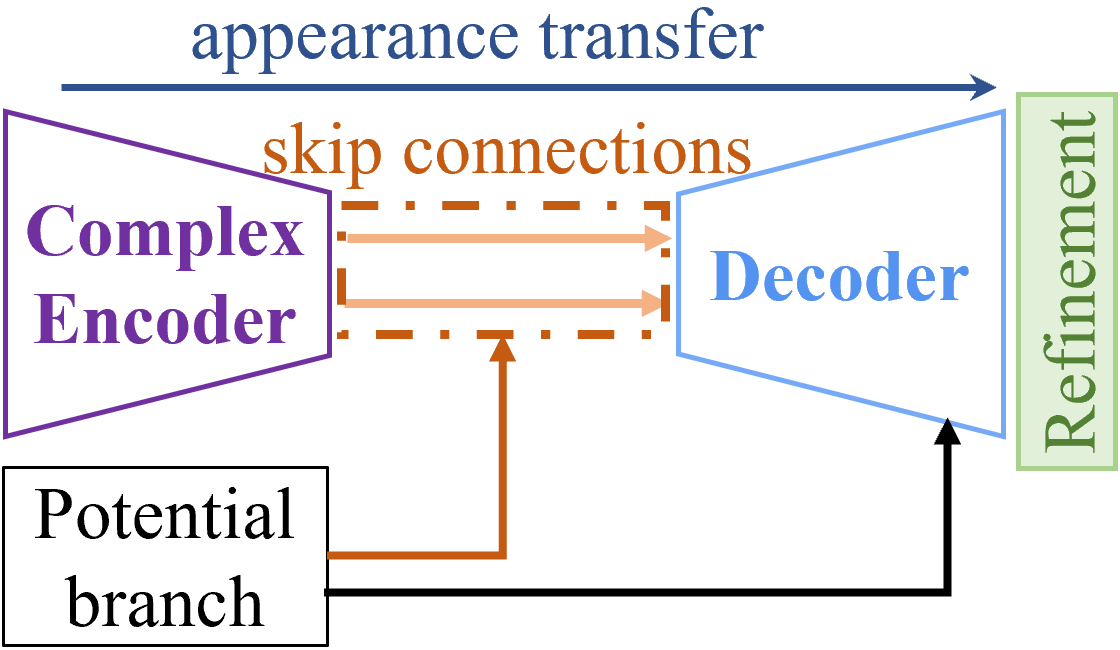} &
			\includegraphics[width=0.46\linewidth]{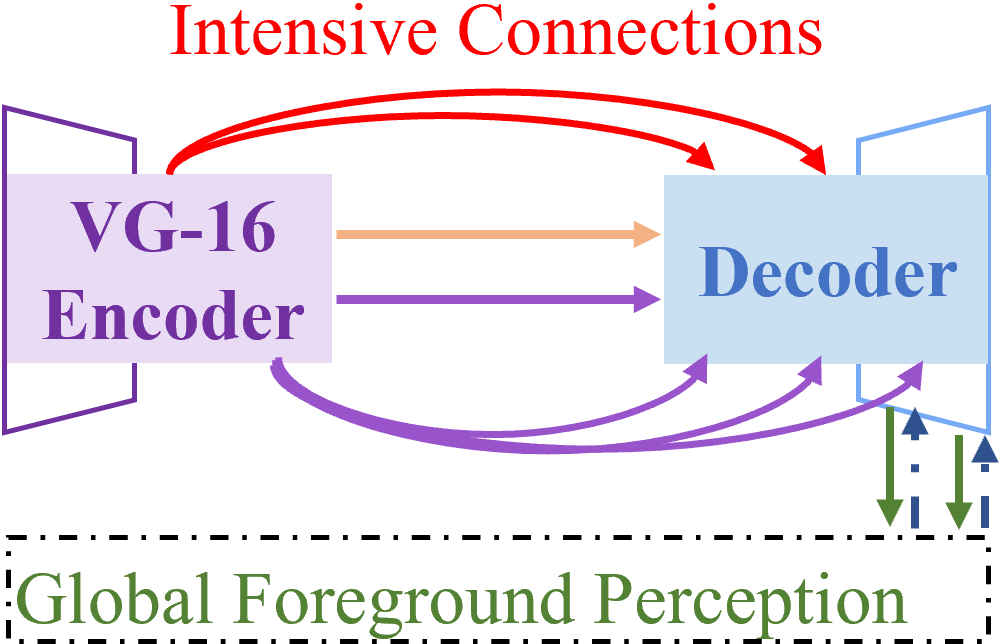} \\
			
			(a) The prevalent pipeline &
			(b) The proposed pipeline \\
		\end{tabular}
	\end{center}
	\vspace{-5mm}
	\caption{The motivation of wider and higher compared to the prevalent pipeline of most existing matting researches~\cite{hou2019context,li2020natural,hao2019indexnet,Tang_2019_CVPR}. We sacrifice model depth for a sustained intermediate resolution (output stride=4) and bridge intensive connections to the decoder. Global foreground perception can contribute to more details. }
	\vspace{-5mm}
	\label{fig:teaser_model}
\end{figure}

Image matting has a wide application in film productions, image editing, live video, etc. Since the first success to explore image matting with convolutional neural networks~\cite{Cho2016Natural}, deep learning has contributed significantly to the performance improvement of alpha mattes. Xu~\et~\cite{Xu2017Deep}combined RGB images with trimaps as conjoint input and proposed an encoder-decoder model to predict alpha mattes, providing primary heuristics for most later researches. Many elaborated models~\cite{Tang_2019_CVPR,cai2019disentangled} are proposed with potential modifications like skip connections, refinement modules, and additional branches to improve the visual and quantitative quality of alpha mattes. These models share an analogous pipeline and a complicated encoder architecture (Fig.~\ref{fig:teaser_model} (a)), the Resnet50 backbone in~\cite{Tang_2019_CVPR}, the Xception-65 encoder in~\cite{hou2019context}, the MobinleNetV2 model in~\cite{hao2019indexnet} etc. Some trimap-free methods~\cite{Qiao_2020_CVPR,Zhang2019CVPR} also exploit deep models to generate alpha mattes. Complex encoders can extract advanced semantics and contribute to the shape completion of foreground objects. However, in this paper, we throw such a doubt: advanced semantics is the decisive element in image matting?

\begin{figure}[t]
	\begin{center}
		\setlength{\tabcolsep}{1pt}{
			\begin{tabular}{ccc}
				\includegraphics[height=0.21\linewidth]{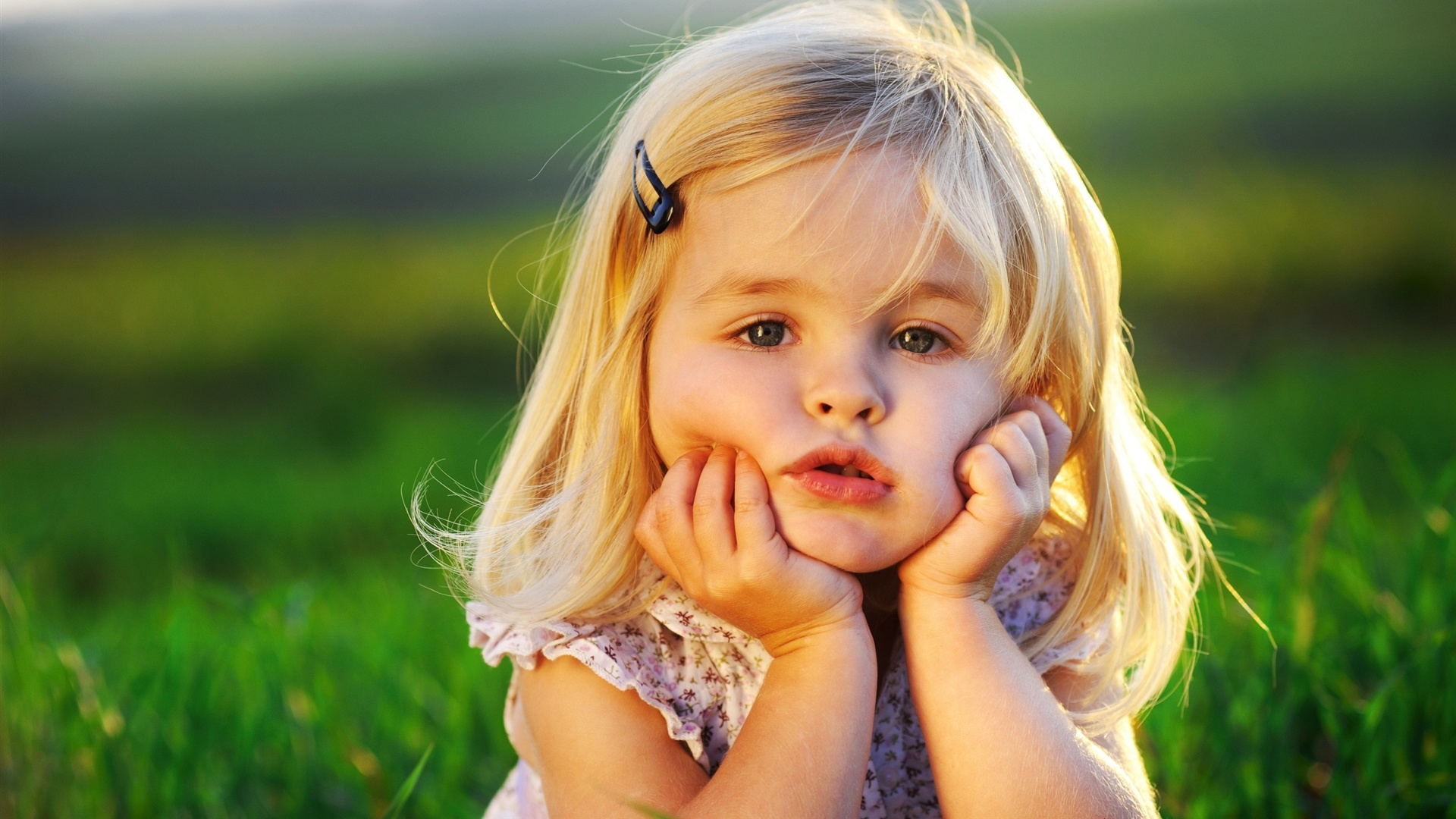} &
				\includegraphics[height=0.21\linewidth]{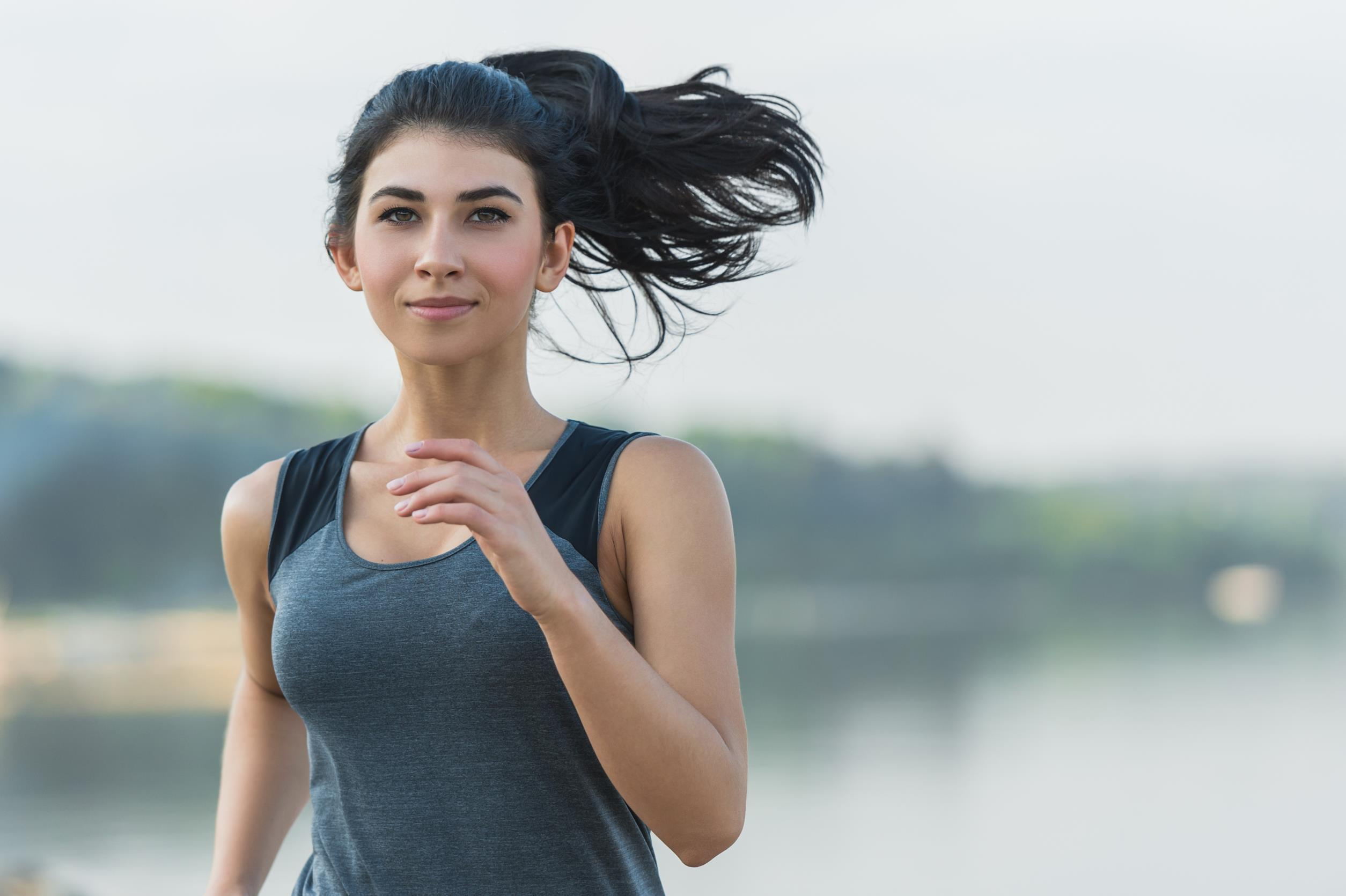} &
				\includegraphics[height=0.21\linewidth]{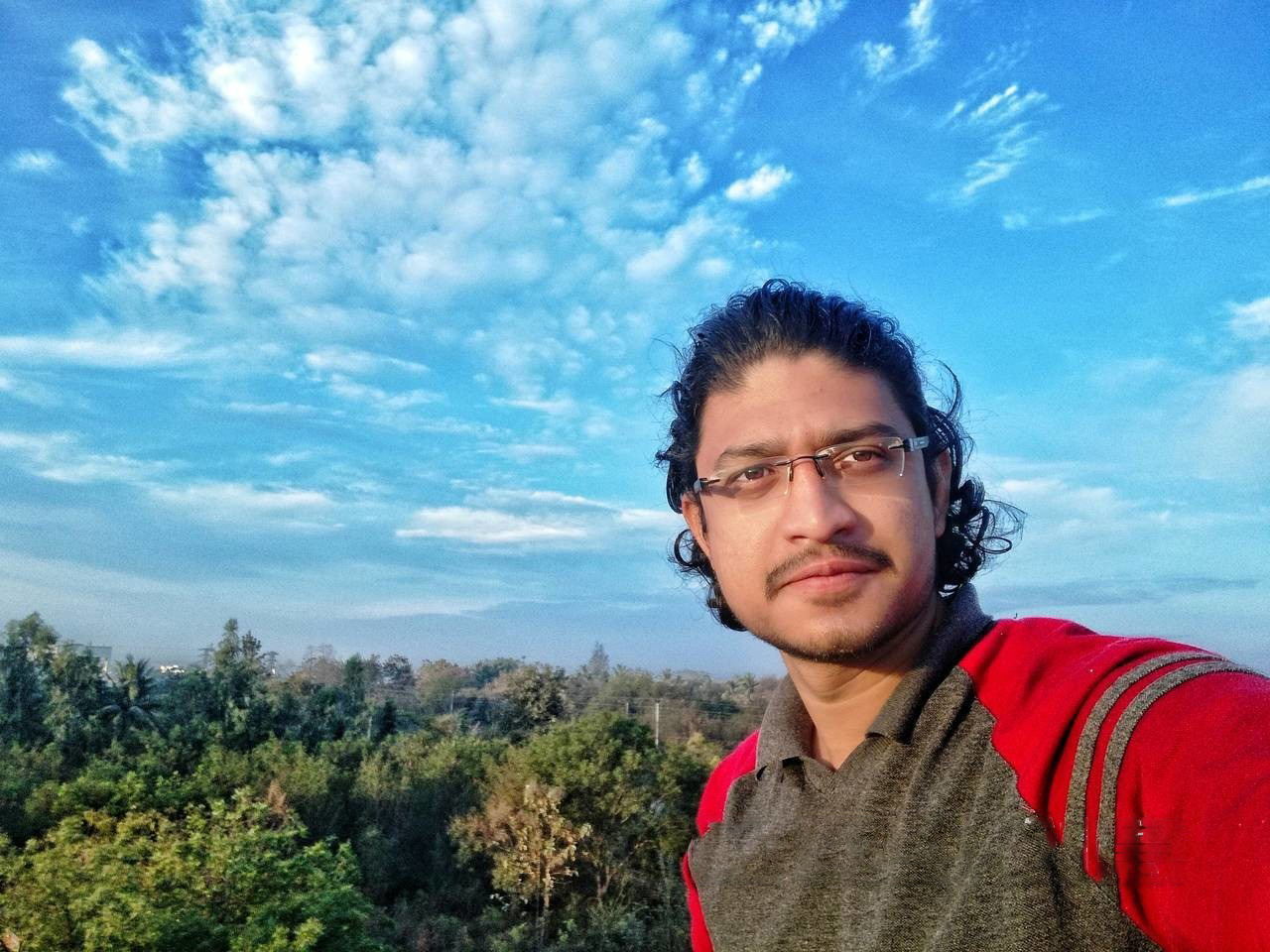} \\
				
				\includegraphics[height=0.21\linewidth]{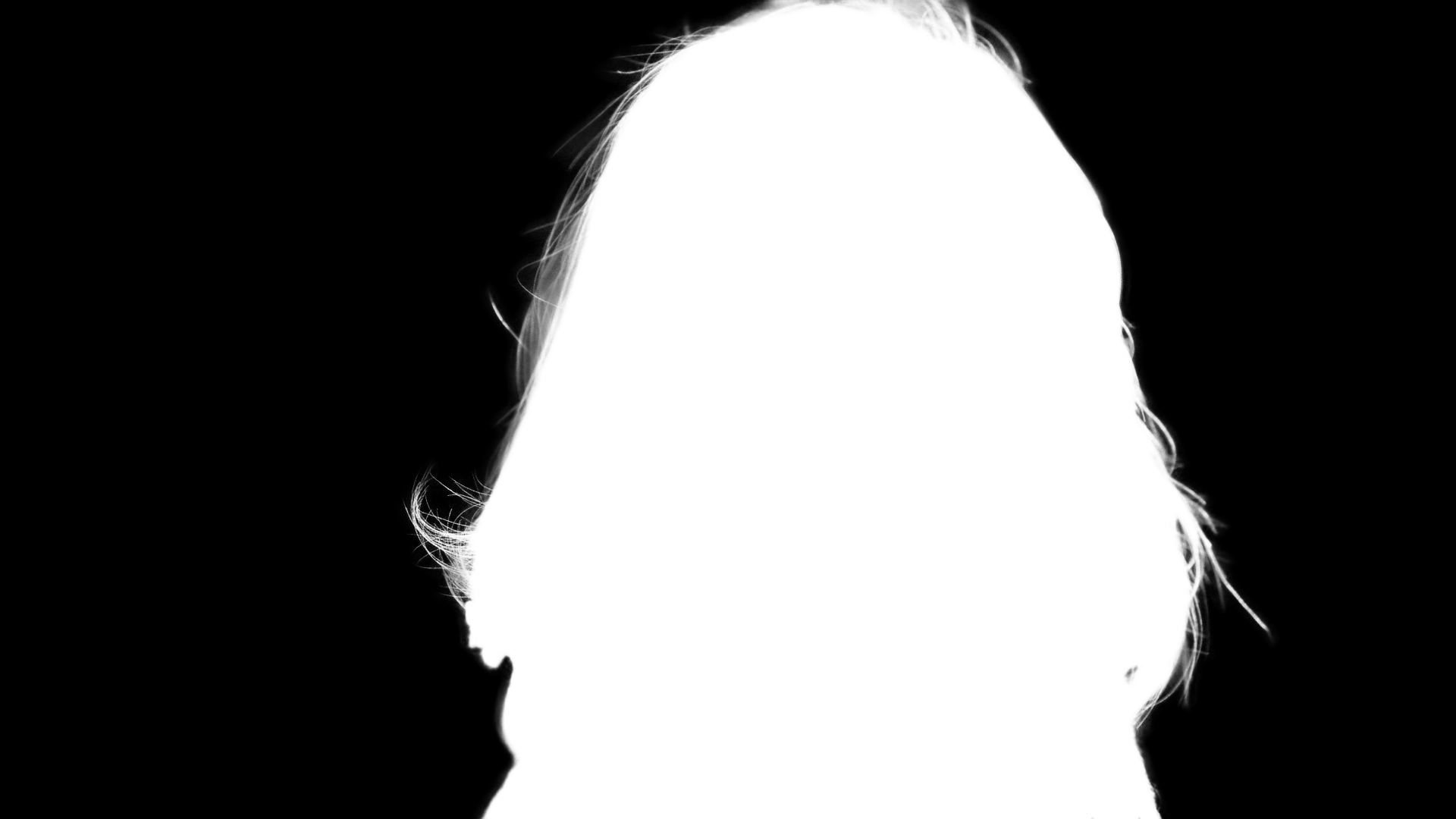} &
				\includegraphics[height=0.21\linewidth]{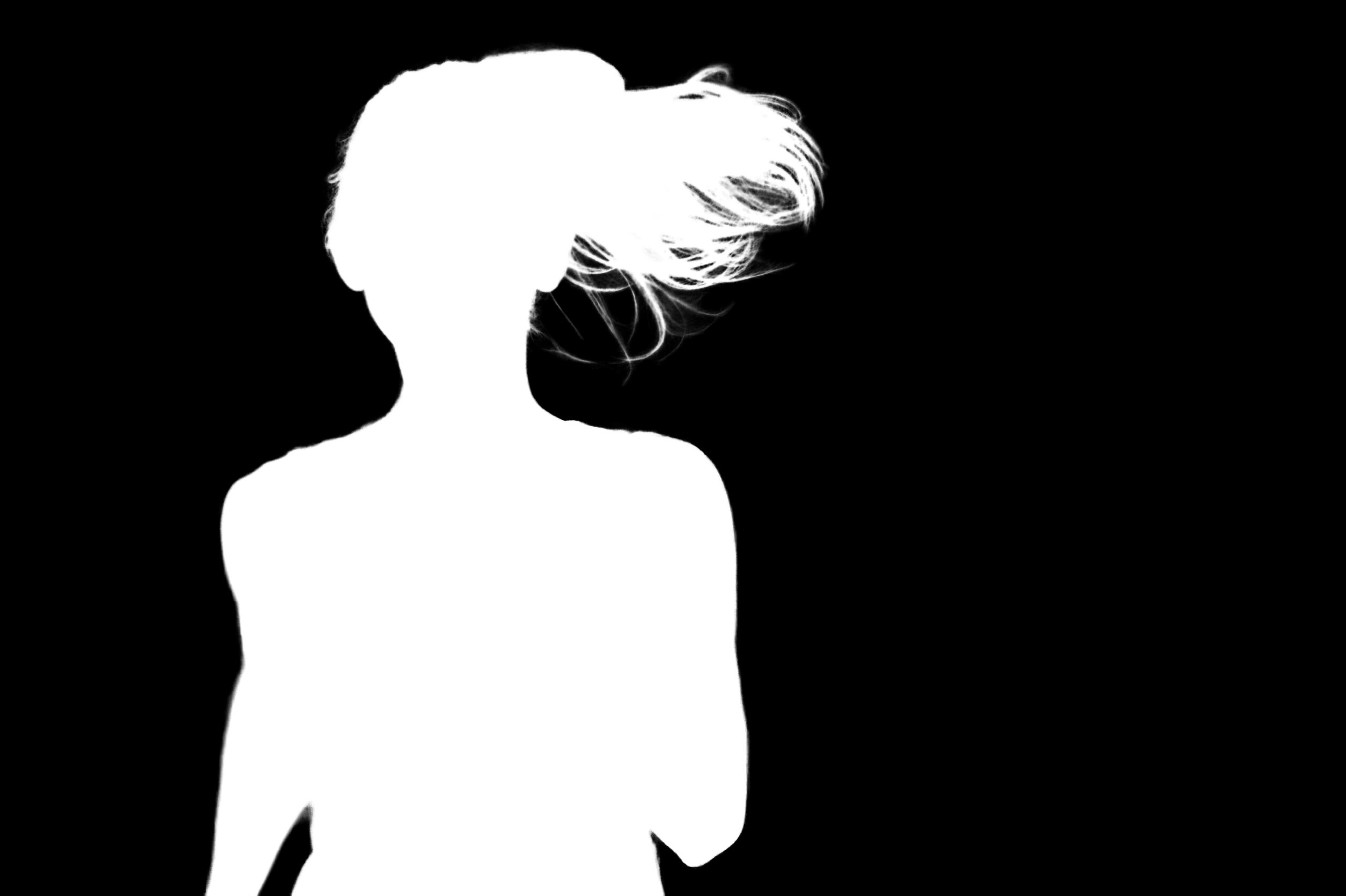} &
				\includegraphics[height=0.21\linewidth]{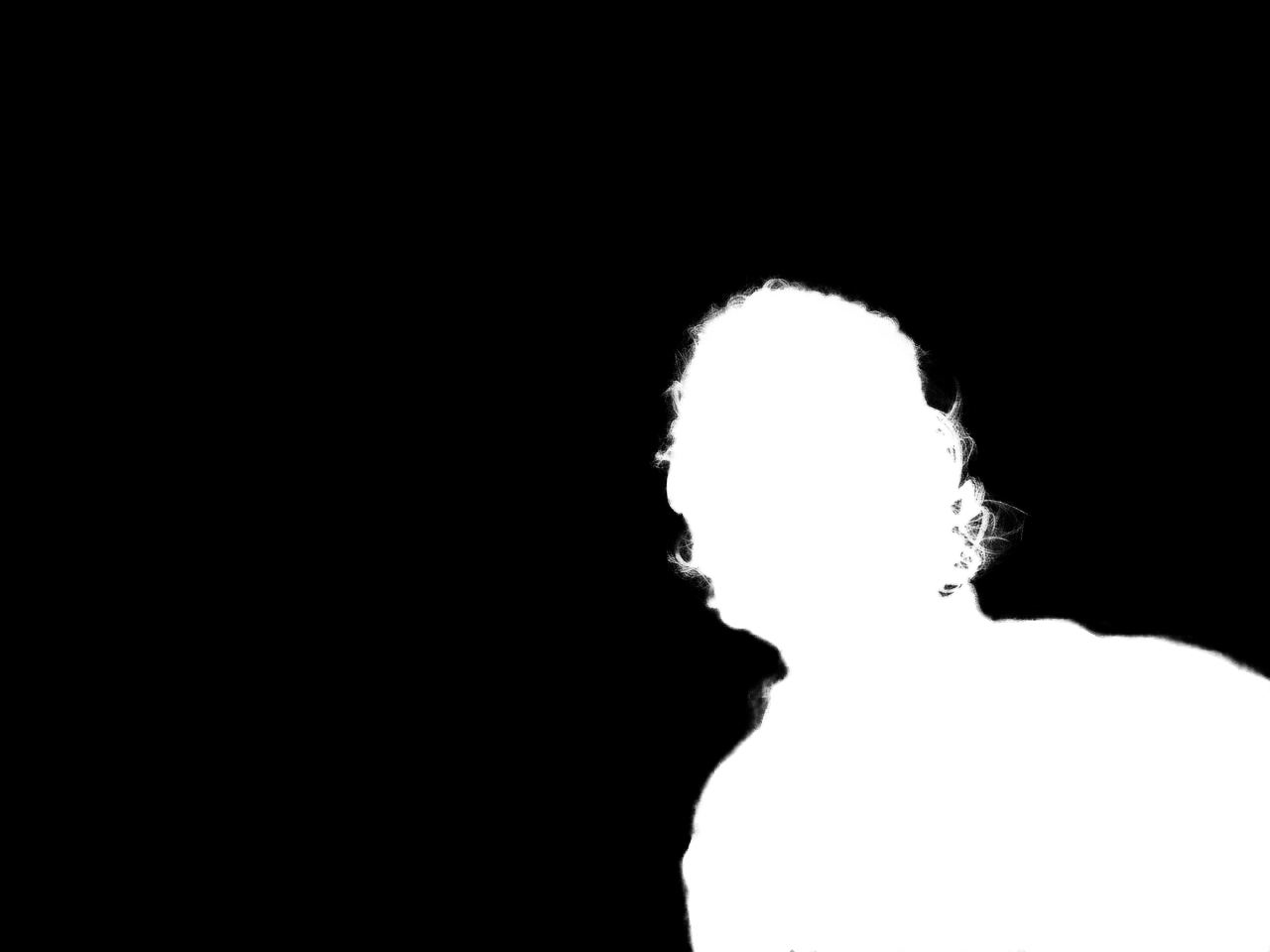} \\
		\end{tabular}}
	\end{center}
	\vspace{-5mm}
	\caption{The alpha mattes produced by our I2GFP on the natural images. }
	\vspace{-5mm}
	\label{fig:teaser}
\end{figure}
The essence of image matting is a pixel-wise estimation, which is a common understanding from the image synthesis equation~\cite{Xu2017Deep}. The anticipated alpha matte should bridge an end-to-end correspondence between the input image and the foreground opacity. This assumption can authorize a great degree of arbitrariness on semantic information (Fig.~\ref{fig:teaser}): a half-length portrait or a head can suggest a human, several leaves or petals can represent a flower. Based on this observation, we can conclude that advanced semantics may not be the consequential factor for image matting. Instead of deep semantics, wider and higher feature representation, perception and communication are more crucial for high-quality alpha mattes.

In this paper, we abandon extremely advanced semantics in favor of intensive integration between various encoder features, and complement the global foreground perception branch to provide a highly expressive appearance. The schematic comparison between our motivation and general existing matting models is illuminated in Fig.~\ref{fig:teaser_model}. Specifically, we adopt a $4$-stride output VGG-16 encoder to retain more input information and bridge intensive connections (IC) to the decoder stage. Each decoder block can catch all low-order layers to integrate wider representation, and the consistent resolution ($output\_stride=4$) can enable such an incremental feature fusion. Simultaneously, we import a global foreground perception (GFP) branch to capture large-field appearance details. Generally speaking, high-resolution feature representation and extensive foreground appearance can maintain more details and textures, and their accumulation can provide wider attribute integration. Therefore, the proposed Intensive Integration and Global Foreground Perception network (I2GFP) can achieve a wider and higher alpha matte estimation. We conduct extensive experiments on public matting datasets, and the state-of-the-art performance can also verify the rationality of our motivation.

Our main contributions are summarized as follows:
\vspace{-2mm}
\begin{itemize}
	\item We propose the Intensive Integration and Global Foreground Perception network (I2GFP), providing a wider and higher feature representation with the sacrifice of model depth. Extensive experiments on public datasets can prove our motivation and the proposed model.
	\item We bridge Interleaved Connections (IC) between the encoder and decoder stage, enabling a wider feature communication.
	\item We present a Global Foreground Perception branch to capture and convert primitive appearance, which can complement rich foreground details. 
\end{itemize}

\vspace{-3mm}
\section{Related Work}
\label{sec:related_work}
\vspace{-1mm}
Here we briefly review traditional matting approaches, then enumerate recent deep-learning-based architectures and analyze their functions and universality.

\textbf{Traditional matting.} Traditional approaches resort to color or texture distribution to estimate alpha mattes. They can be divided into two categories according to different ways of uncertainty expansion: sampling-based and affinity-based methods. Sampling-based solutions~\cite{gastal2010shared} connect the pixel-wise uncertainty in foreground and background to infer the labels of transition regions. As for affinity-based methods~\cite{Levin2007A}, they calculate the correlations between certain and transition regions, then propagate the labels to the whole image. Potential certainty is provided by trimap or scribbles priors, and both involve user interactions to indicate some label attributes. Each pixel in trimap shares an explicit label correspondence, white-foreground, black-background and gray-transition. While scribbles have limited prior annotations suggested by several scribbles to represent foreground and background. Due to the restricted expressive ability of hand-crafted features, traditional matting approaches usually generate ambiguous alpha mattes when the input images have complex colors or textures.

\textbf{Deep-learning-based matting.} Recent deep learning models can achieve comparable results benefiting from the structure representation of powerful convolution neural networks. Generally speaking, existing matting architectures have two solutions, with trimaps as confine and end-to-end automatic models. Xu~\et~\cite{Xu2017Deep} employ concatenated RGB images and trimaps as input, regressing alpha mattes with an encoder-decoder model. Most later methods~\cite{cai2019disentangled,hou2019context,Tang_2019_CVPR} follow this to design their models, using a complex encoder to extract advanced semantics and achieve original resolution with a U-Net like decoder. Hao~\et~\cite{hao2019indexnet} and Dai~\cite{dai2021learning} unify upsampling operators with the index functions to improve encoder-decoder network. Li~\et~\cite{li2020natural} utilize a guided contextual attention to improve the transmission of the opacity. Though the trimap expansions in SIM~\cite{sun2021semantic} can extend the range of transition capture, they require alpha categories.

Several researches~\cite{Qiao_2020_CVPR,Qiao2020MSIA,Yu_2021_ICCV,Zhang2019CVPR} achieve trimap-free architectures, but they have limited generalization in natural images and also rely on complex encoders to extract extremely advanced semantics. There are also some methods relying on multiple stages or feedback to perform model optimization. \cite{Cho2016Natural} fuses trimap labeling and image matting as a two-phase framework. Interactions-based methods~\cite{Yang2018Active,wei2021improved} resort to user feedback to refine alpha mattes. Some approaches~\cite{yu2021mask} and~\cite{lin2021real} can produce results with additional guidance like masks, backgrounds. % It is well known that the essence of image matting is the precise estimation of pixel-wise opacity. Therefore, we argue that extremely advanced semantics are not the only consequential factor for image matting. We resort to VGG-16~\cite{Karen2015VGG} to extract semantics and employ interleaved connections (\emph{IC}) and global foreground perception (GFP) to achieve wider and higher feature representation.
\begin{figure*}[t]
	\begin{center}
		\includegraphics[width=\linewidth]{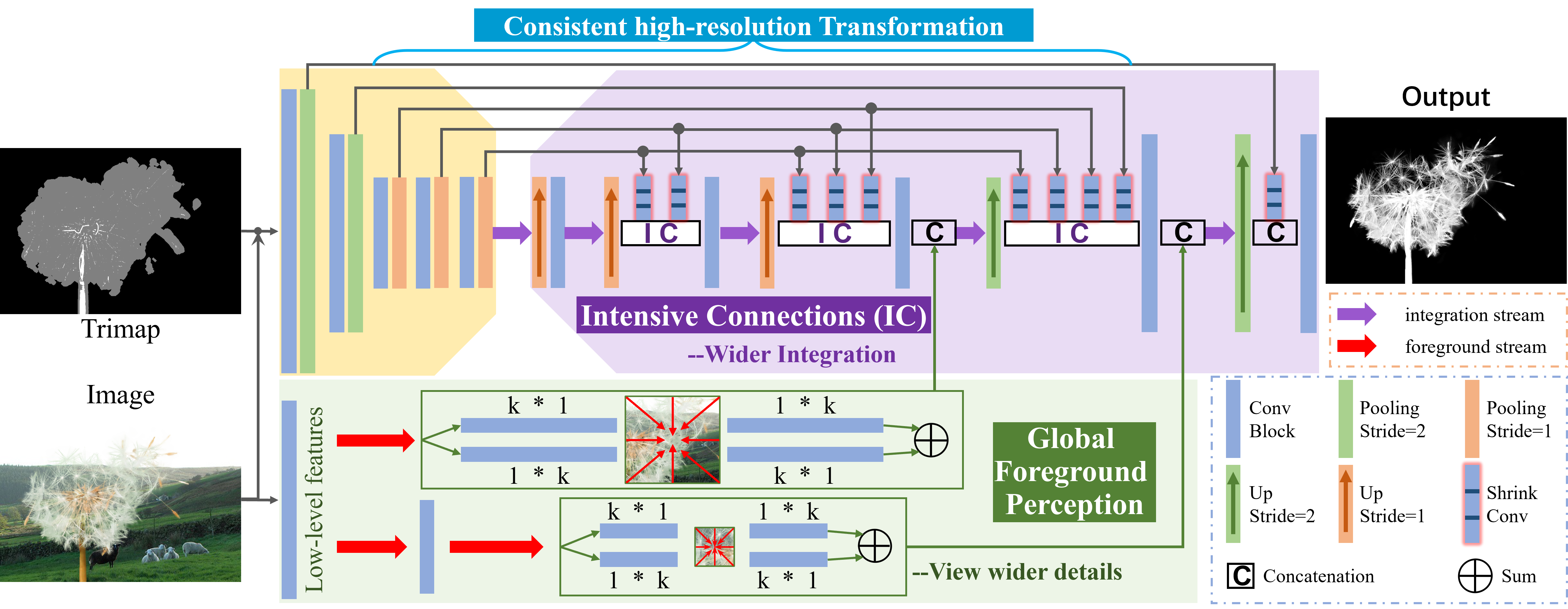}
	\end{center}
	\vspace{-5mm}
	\caption{The overall architecture of the proposed Intensive Integration and Global Foreground Perception network (I2GFP). We employ simple backbone to extract necessary semantics and utilize intensive connections to integrate different-level features. The global foreground perception can capture rich appearances to complement details. }
	\vspace{-5mm}
	\label{fig:pipeline}
\end{figure*}

\vspace{-3mm}
\section{Methodology}
\label{sec:method}
\vspace{-2mm}
Our motivation is to provide wider and higher feature fields for alpha estimation. Although model depth can contribute to advanced semantics with an accumulated receptive field, we argue that extreme semantics are not consequential. On the one hand, the foreground classes enjoy a high granularity of freedom in natural matting images and bring huge ambiguity for vanilla semantics. It could be anything sharing a transparent uncertain region; even part of them can also be the target. On the other hand, image matting requires pixel-wise opacity and all alpha values expect a wider feature range and higher field to receive more opacity perception. The wider range can combine different-level feature representation, while the higher field can retain extensive appearances and details to enrich fine-grained prediction. Based on the above observations, we propose the Intensive Integration and Global Foreground Perception network (I2GFP), with intensive connections (IC) and global foreground perception (GFP), to achieve wider and higher feature representation. In this section, we first introduce the whole architecture~\ref{ssec:network}, then the IC module~\ref{ssec:ic} and GFP branch~\ref{ssec:gfp}.

\vspace{-2mm}
\subsection{Network Architecture}
\label{ssec:network}
\vspace{-1mm}
The architecture of the Intensive Integration and Global Foreground Perception network (I2GFP) is illuminated in Fig.~\ref{fig:pipeline}. We take an RGB image and trimap as concatenated input ($\mathbb{R}^{4 \times H \times W}$). Trimap is imported to suggest transition regions. To fully demonstrate the validity of I2 and GFP in capturing matting features, we harness original VGG-16 to capture necessary semantics and record the features from different blocks as $\mathcal F_{E}^{i} \in \mathbb{R}^{c \times h \times w}$, where $i$, $c$, $h$, and $w$ denote the block index, channels, height, and width of features. We only reserve the stride=2 pooling operation in the first two blocks, thus can generate a high-resolution intermediate feature $\mathcal F_{E}^{i} \in \mathbb{R}^{c \times H/4 \times W/4}, i \in \{3,4,5\}$. This $output\_stride=4$ design can extract sufficient semantics and preserve more initial attributes.

For the decoder stage, we adopt Intensive Connections (IC) to obtain a wider feature representation. We also propose a Global Foreground Perception (GFP) branch to provide necessary appearances. For a better training procedure, we incrementally add the IC and GFP module to achieve optimal convergence.

\vspace{-2mm}
\subsection{Intensive Connections}
\label{ssec:ic}
\vspace{-1mm}
Although common skip connections can bridge the feature transport, each decoder layer can only observe the corresponding encoded features, resulting in direct calculation between skip connections and upsampled features: $\mathcal F_{D}^{i} = \Phi(\theta) \sim \mathcal Cat \lbrace \mathcal F_{E}^{i}, \mathcal F_{U}^{i-1} \rbrace$, where $\mathcal F_{D}^{i} \in \mathbb{R}^{c \times h \times w}$ and $\mathcal F_{U}^{i} \in \mathbb{R}^{c \times h \times w}$ represent the $i$-layer encoder feature and $i$-layer unsampling feature. Such connection ignores the context relevance desired by image matting. Considering the example in Fig.~\ref{fig:pipeline}, the stems, branches, and catkins of the dandelion correspond to a potentially different range of receptive fields, but the correlations and opacity between them are mutually enforced by each other. Therefore, we employ intensive connections (IC) to improve this correlation among different-level features.

Formally speaking, we define the calculation of step-wise convolution in decoder stage as follows:
\begin{equation}
\vspace{-2mm}
\mathcal F_{D}^{i} = \Phi(\theta) \sim \mathcal Cat \lbrace \mathcal F_{E}^{5}, \mathcal F_{E}^{4} ... \mathcal F_{E}^{i}, \mathcal F_{U}^{i} \rbrace,
\vspace{-2mm}
\end{equation}
except for the output convolution, each decoder layer takes the last upsampling feature and all low-order encoder features as input to obtain wider feature representation. For example, the fourth decoder layer has a concatenated input of $\mathcal F_{E}^{5}$, $\mathcal F_{E}^{4}$ and $\mathcal F_{U}^{4}$, and output $\mathcal F_{D}^{4}$. IC can promote the communication between different receptive fields. We use shrink convolution to transport the encoder features with a channels of $16$. The final output layer only takes $\mathcal F_{U}^{1}$ and $\mathcal F_{E}^{1}$ as input to preserve more primitive foreground appearance.

Compared to previous dense connections~\cite{Zhang2021MMM,Zhang2021TIP}, IC has two novel parts for image matting. The first starts with the third encoding layer and ends with the third decoding layer, sharing consistent resolution, and we integrate them by shrinking convolution and the concatenation operation. The second part is the combination between the decoder features and the low-level information from GFP. The consistent high-resolution integration can promote context-related foreground communication, and the combination with primitive textures can provide more foreground details for alpha mattes.

\vspace{-2mm}
\subsection{Global Foreground Perception}
\label{ssec:gfp}
\vspace{-1mm}
Many matting works~\cite{dai2021learning,Xu2017Deep} combine the initial encoded features in the back-end layers of the decoder to supplement the foreground details. The deep network can extract abstract semantics, while the low-level features contain more primitive textures (appearances and details). Inspired by the large kernels in~\cite{peng2017large}, we propose our global foreground perception (GFP) branch to capture low-level details. Given the RGB input, we first perform two downsampling convolutions to reduce calculation space. Then two global convolutional layers can perceive large-field foreground appearances with a large kernel size $k$. Specially, we employ a combination of $1 \times k + k \times 1$ and $k \times 1 + 1 \times k$ convolutions like~\cite{peng2017large}, enabling more feature perception and integration in a large $k \times k$ region. $k$ equals space size is the ideal receptive field for capturing the full foreground, but we use half the spatial size comprising the execution cost. GFP can perceive sufficient foreground appearances and details, and the large-kernel global convolutions can retain the contextual opacity variations within the RGB input.

Compared to Gloabl Convolutional Network~\cite{peng2017large}, there are two main innovations of GFP. 1) We only use large kernels on the initial image features to extract foreground details and bridge the direct transformation between initial image features and alpha mattes. 2) Compared to the semantic problems, image matting requires much more foreground textures and boundaries from the input image. The large kernels in the two global convolutional layers of GFP are 255 and 127. Both are the largest size considering the GPU capacity. Small kernels like 3,5...15 can also extract image features, but they are
separated from the requirements of alpha mattes for as many foreground details.

\vspace{-2mm}
\subsection{Loss Functions and Implementation Details}
\label{ssec:loss_details}
\vspace{-1mm}
\textbf{Loss functions.} We use $\mathcal L_{1}$ loss, composition loss ($\mathcal L_{comp}$) and Laplacian loss ($\mathcal L_{lap}$) to ensure the pixel-wise accuracy, and $\mathcal L_{grad}$ loss to balance the gradient.
\vspace{-3mm}
\begin{flalign}
\label{eq:L1_loss}
&\qquad\quad \mathcal L_{1}=\sum_{j}|\alpha^{j}_{p}-\alpha^{j}_{g}|,&
\end{flalign}
\vspace{-6mm}
\begin{flalign}
\label{eq:comp_loss}
&\qquad\quad \mathcal L_{comp}=\sum_{j}|C^{j}_{g}-\alpha^{j}_{p}F^{j}_{g}-(1-\alpha^{j}_{p})B^{j}_{g}|,&
\end{flalign}
\vspace{-6mm}
\begin{flalign}
\label{eq:grad_loss}
&\qquad\quad \mathcal L_{grad}=\sum_{j}|\bigtriangledown\alpha^{j}_{p}-\bigtriangledown\alpha^{j}_{g}|,&
\end{flalign}
\vspace{-6mm}
\begin{flalign}
\label{eq:Lap_loss}
&\qquad\quad \mathcal L_{lap}=\sum^{5}_{s=1}2^{s-1}|L^{s}_{py}\alpha_{p}-L^{2}_{py}\alpha_{g}|,&
\end{flalign}
\vspace{-6mm}
\begin{flalign}
\label{eq:total_loss}
&\qquad\quad \mathcal L_{total} = \mathcal L_{1} + \mathcal L_{comp} + \mathcal L_{grad} + \mathcal L_{lap}.&
\end{flalign}
where $\alpha^{j}_{p}$ and $\alpha^{j}_{g}$ correspond to the alpha values at pixel $j$ inside unknown regions, $p$ and $g$ denote the predicted and ground truth alpha mattes, $\alpha^{j}_{p},\alpha^{j}_{g} \in [0,1]$. $L^{s}_{py}$ represents the Laplacian functions. And the final loss function is the direct linear combination of $\mathcal L_{1}$, $\mathcal L_{comp}$, $\mathcal L_{grad}$, and $\mathcal L_{lap}$.

\noindent\textbf{Implementation details.} We employ the PyTorch deep learning framework to implement I2GFP, and the training environment is $4$ Tesla V100 graphics cards. We first train the encoder-decoder integration model for $200,000$ iterations with a batch size of $10$ and an initial learning rate of $4e^{-4}$. The VGG-16 backbone is loaded from the pre-trained model, and other parameters are randomly initialized from a Gaussian distribution. All training images are randomly cropped to 512 $\times$ 512, 640 $\times$ 640 or 800 $\times$ 800, then resized 512 $\times$ 512, thus the large kernels in the two global convolutional layers of GFP are $255$ and $127$, respectively. We also employ horizontal random flipping, jitter and affine translation to make data augmentation. We use the Adam optimizer to optimize the network and adopt the Cosine Annealing strategy to schedule the learning rate. We then add the GFP branch to train the whole model with the same hyperparameters. The first training stage takes three days, and the full I2GFP takes additional two days to achieve the best results. 

\vspace{-3mm}
\section{Experiments}
\label{sec:exp}
\vspace{-2mm}
In this section, we evaluate the proposed I2GFP on the public matting datasets and natural images. The comparisons involve visual and quantitative demonstrations with the SOTA approaches and ablation study. The four evaluation metrics include the summation of absolute differences (SAD), mean square error (MSE), the gradient (Grad), and connectivity (Conn) proposed by~\cite{rhemann2009perceptually}. Considering the fairness, most SOTA methods in our experiments are trimap-based. Unless otherwise noted, all metrics are calculated inside the transition regions--the lower the values of all evaluation metrics, the better the predicted alpha mattes. % More visual comparisons can refer to the supplementary materials. 

\begin{figure*}[t]
	\begin{center}
		\setlength{\tabcolsep}{0.6pt}{
			\begin{tabular}{cccccc}			
				\includegraphics[width=0.16\linewidth]{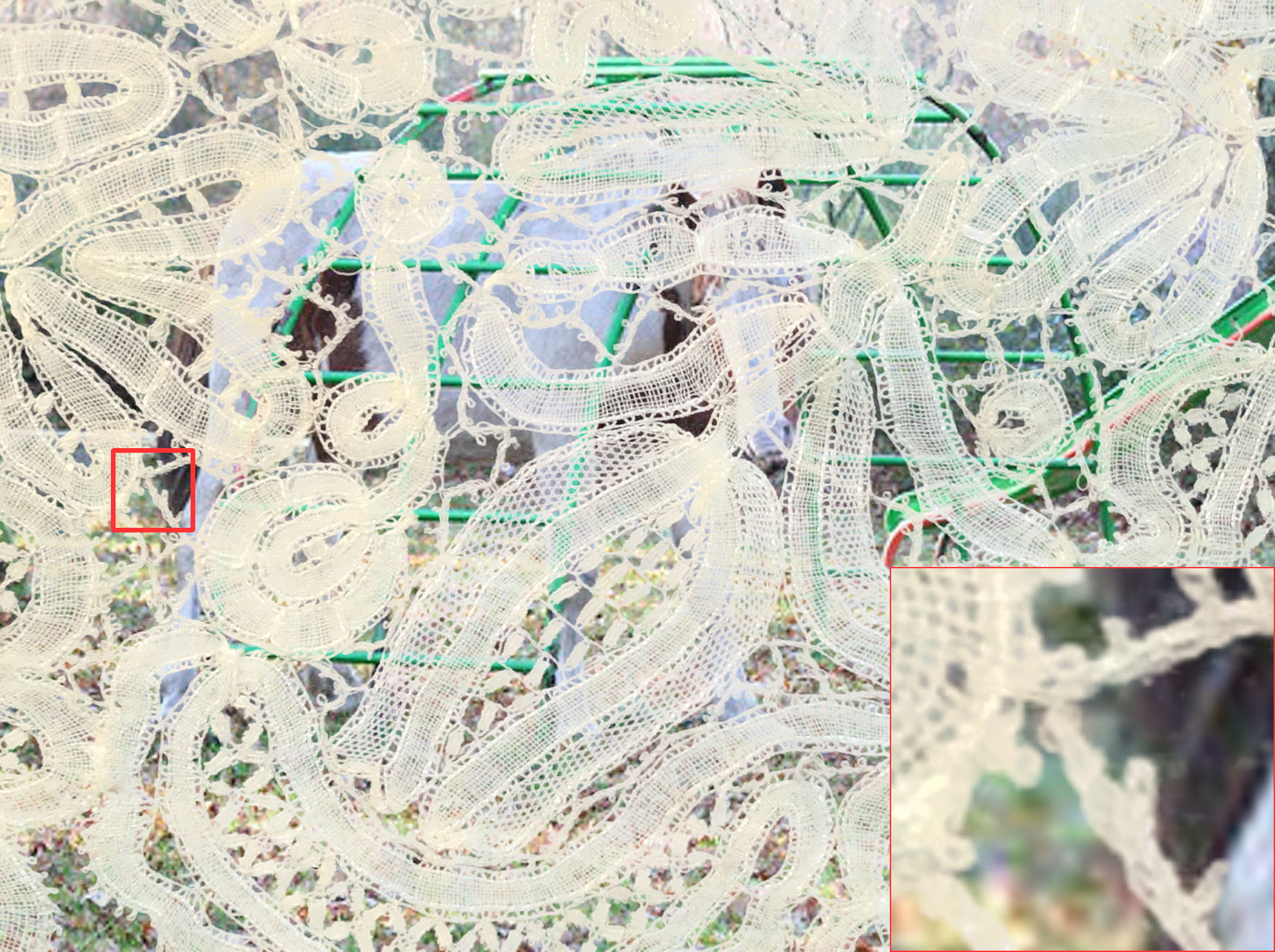} &
				\includegraphics[width=0.16\linewidth]{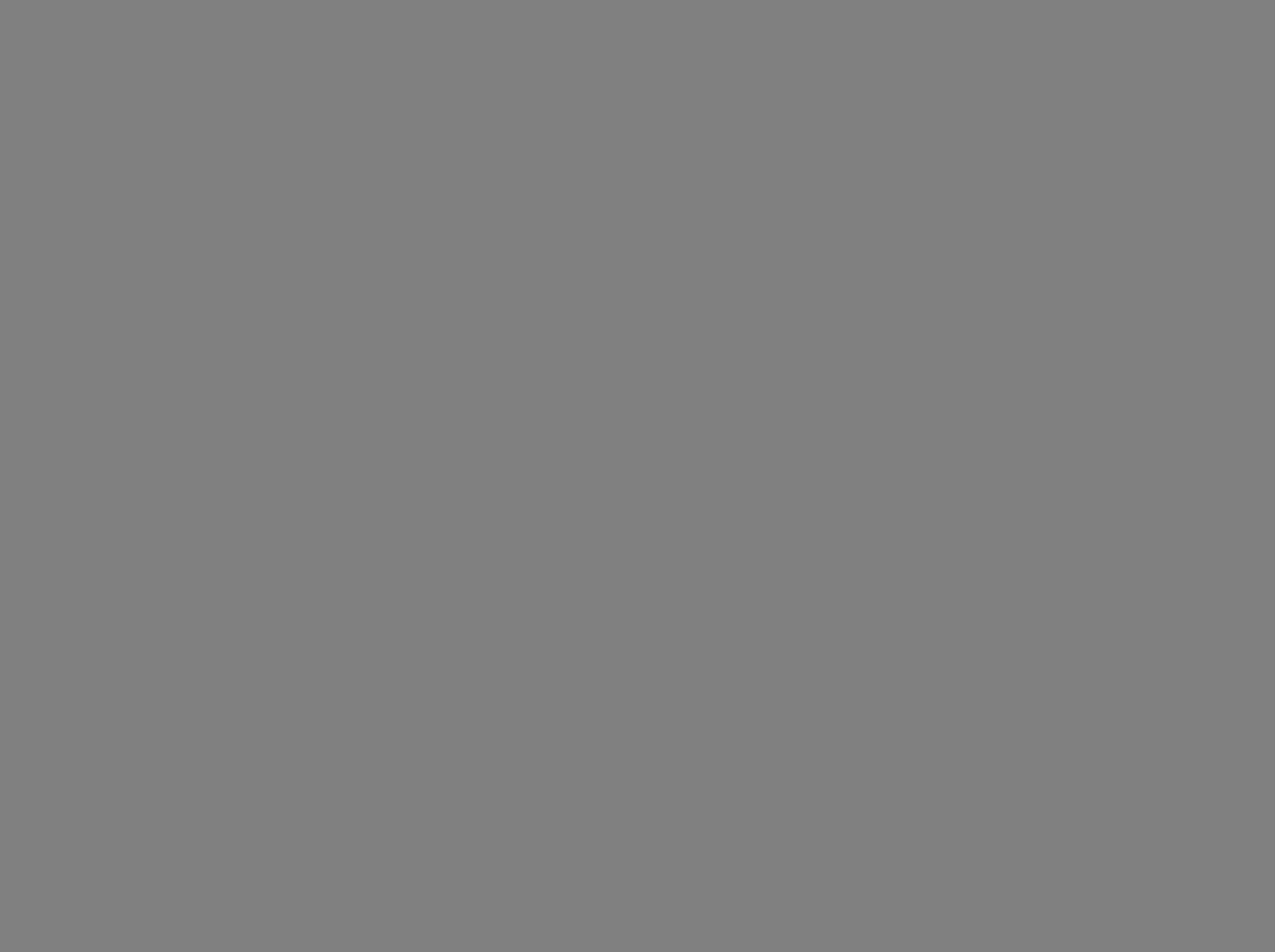} &
				\includegraphics[width=0.16\linewidth]{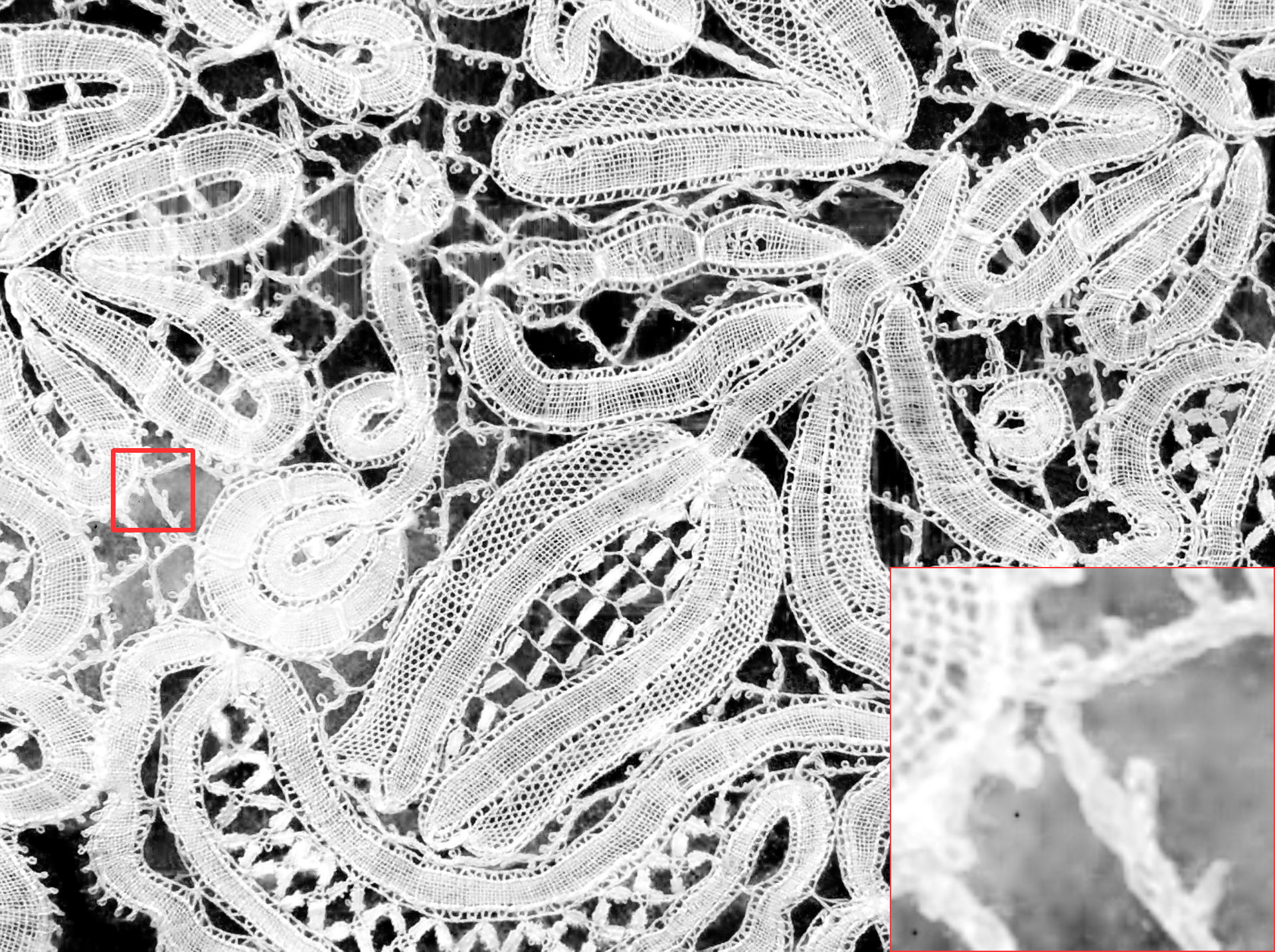} &
				\includegraphics[width=0.16\linewidth]{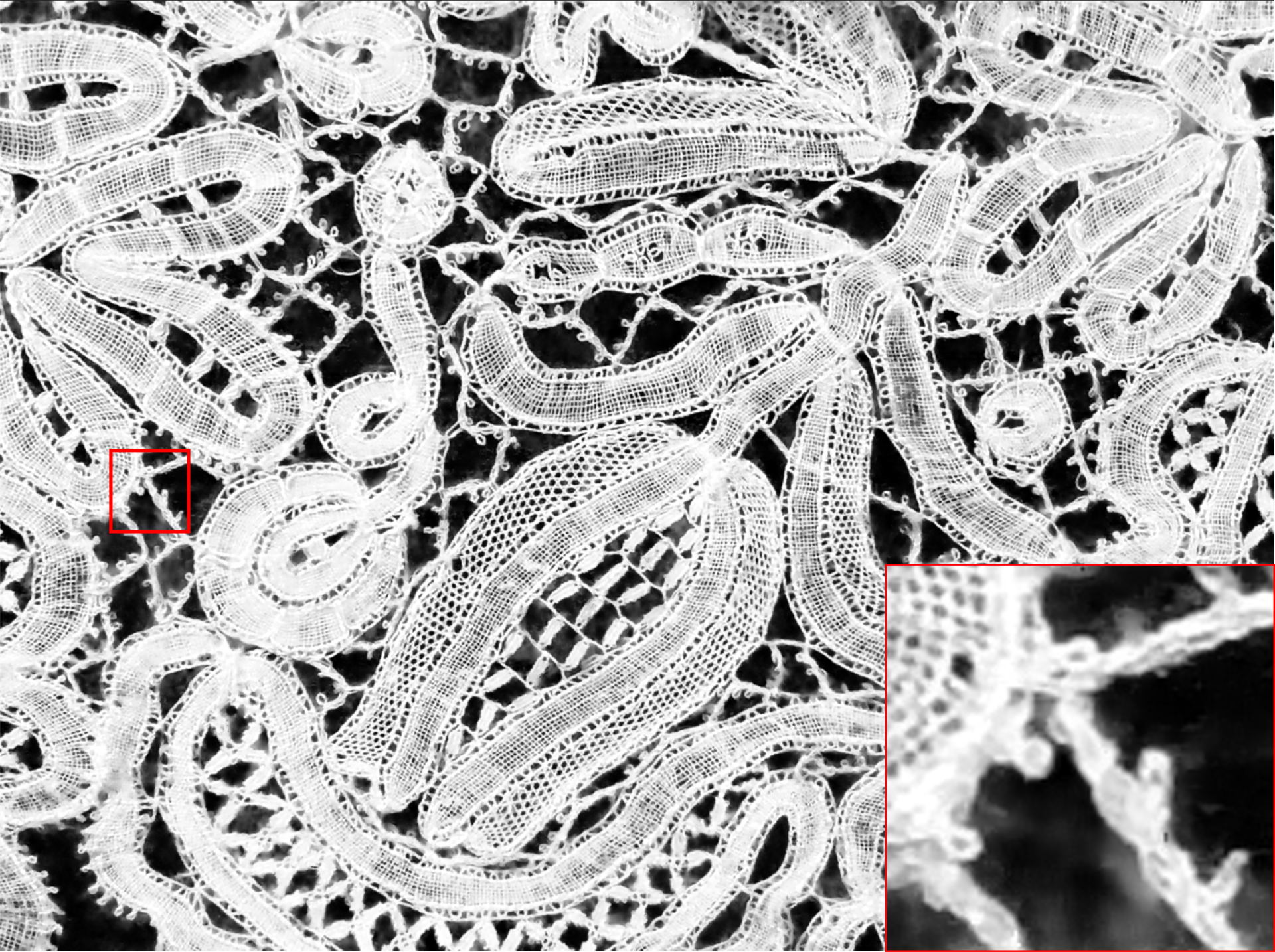} &
				\includegraphics[width=0.16\linewidth]{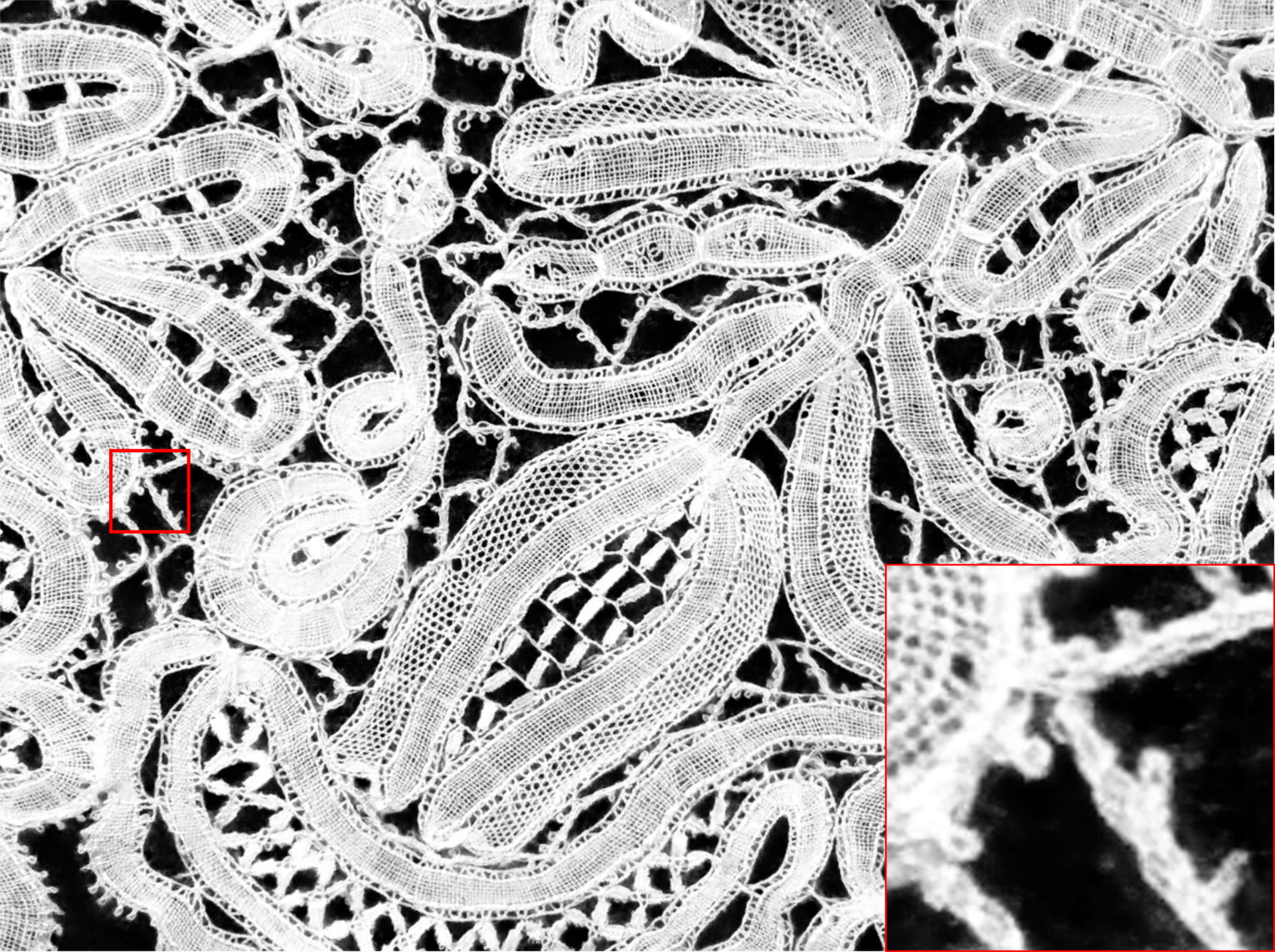} &
				\includegraphics[width=0.16\linewidth]{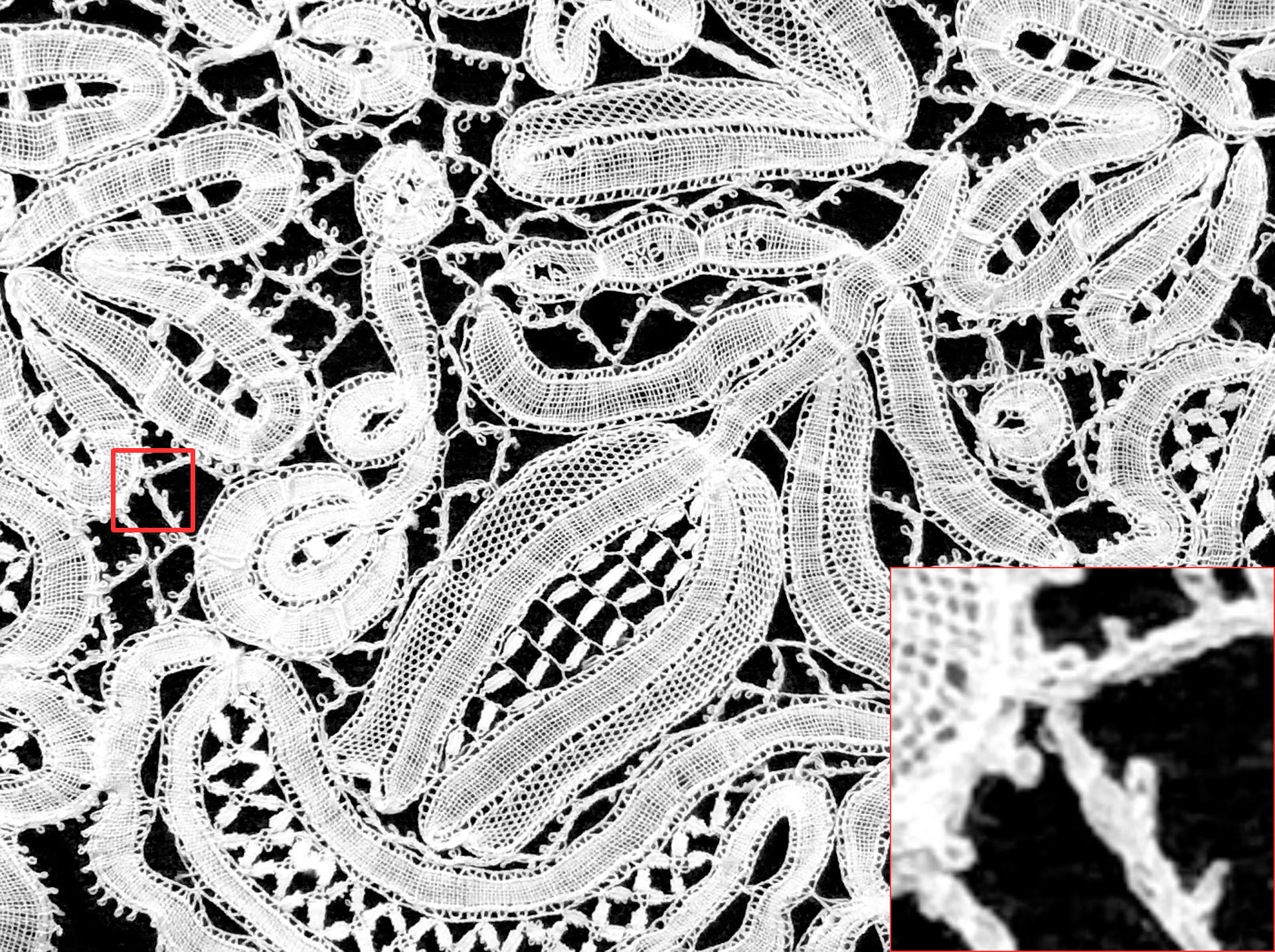} \\
				
				\includegraphics[width=0.16\linewidth]{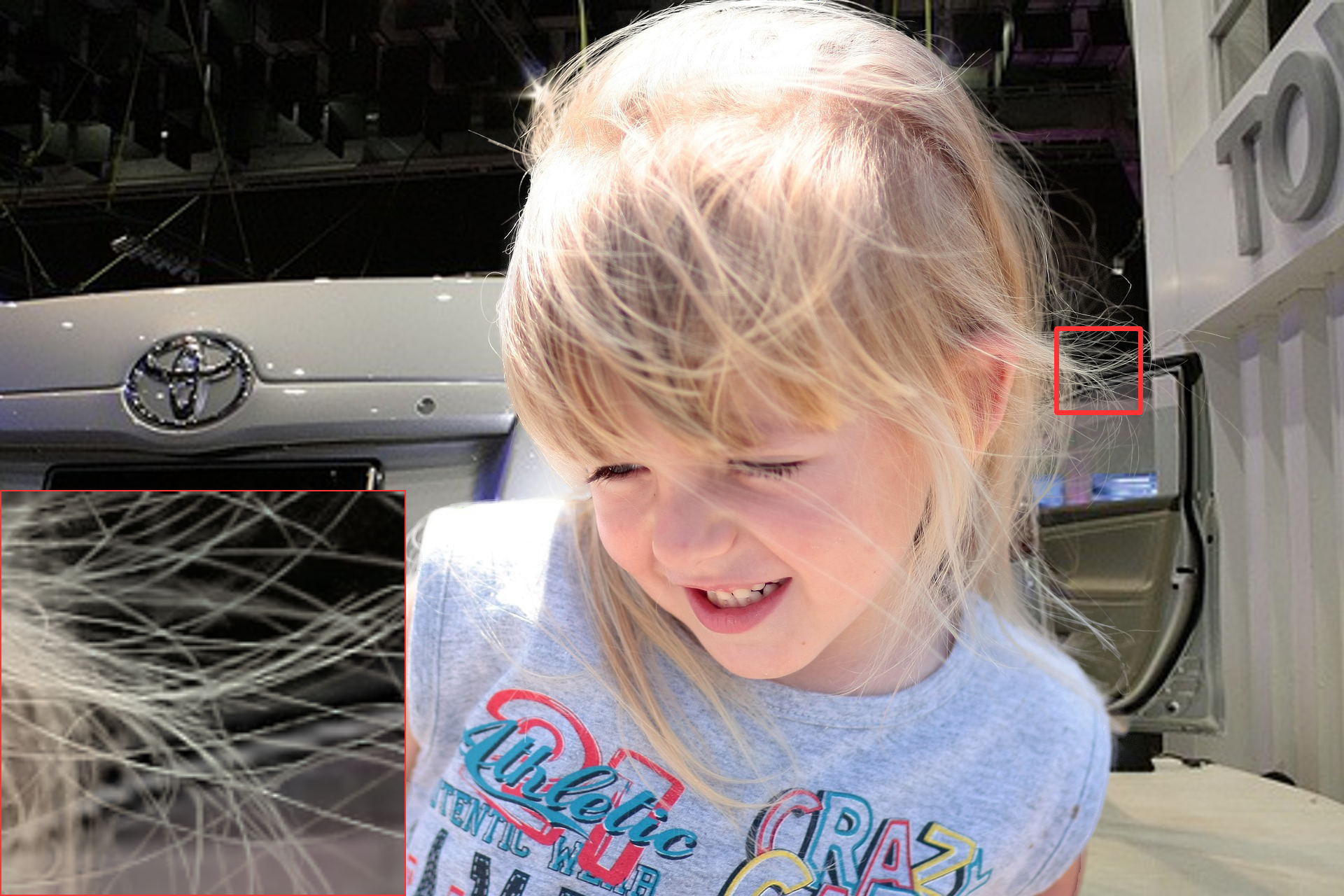} &
				\includegraphics[width=0.16\linewidth]{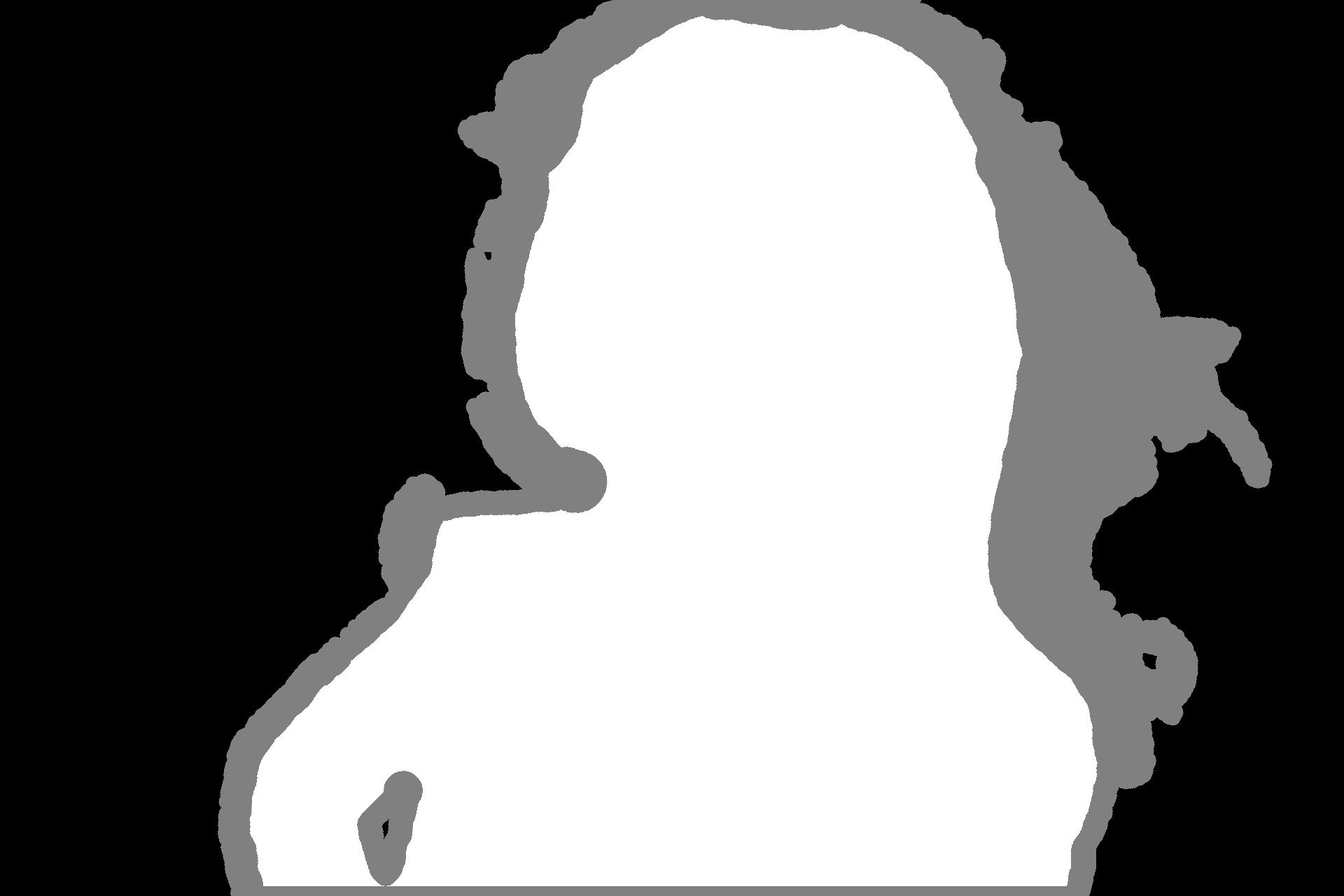} &
				\includegraphics[width=0.16\linewidth]{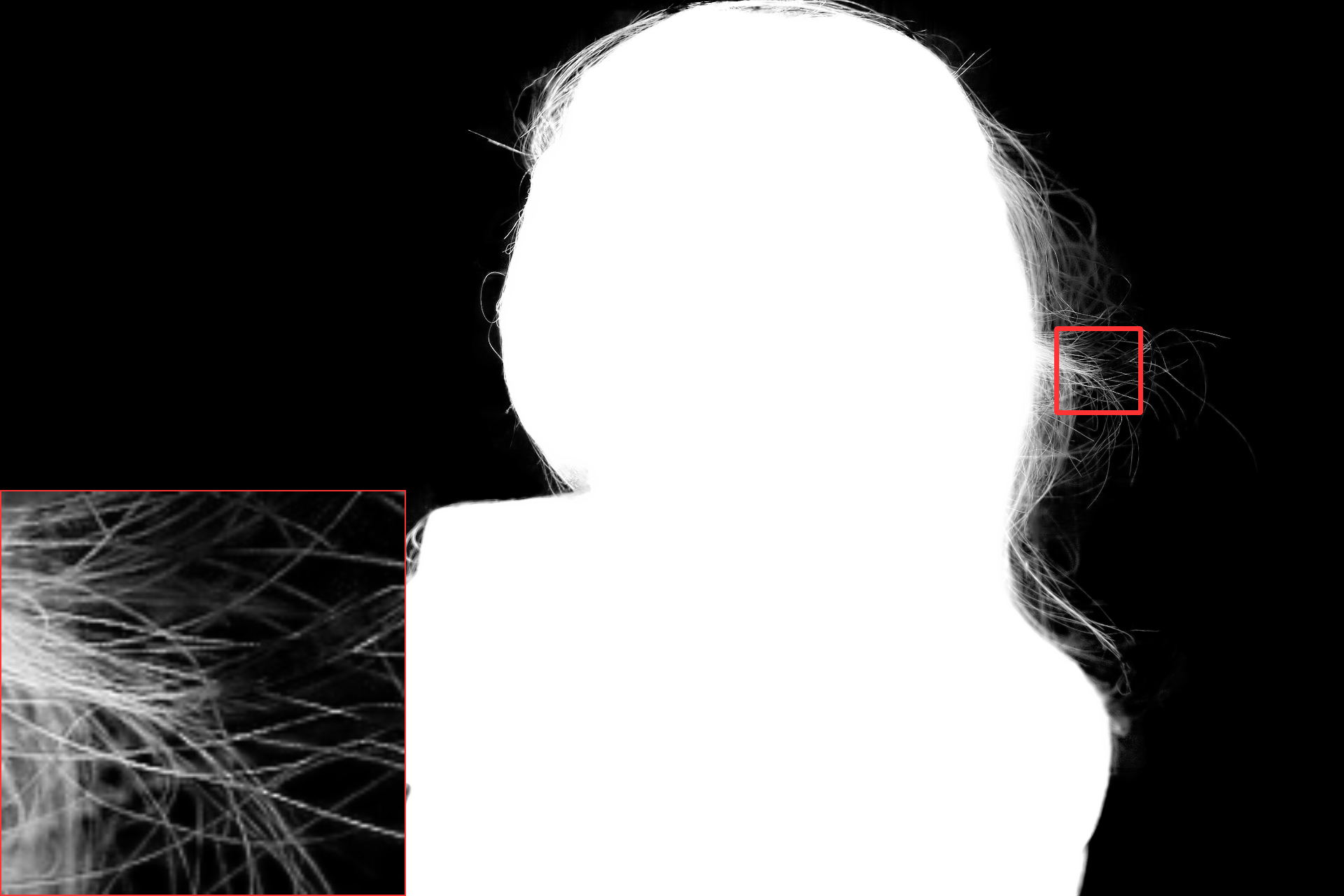} &
				\includegraphics[width=0.16\linewidth]{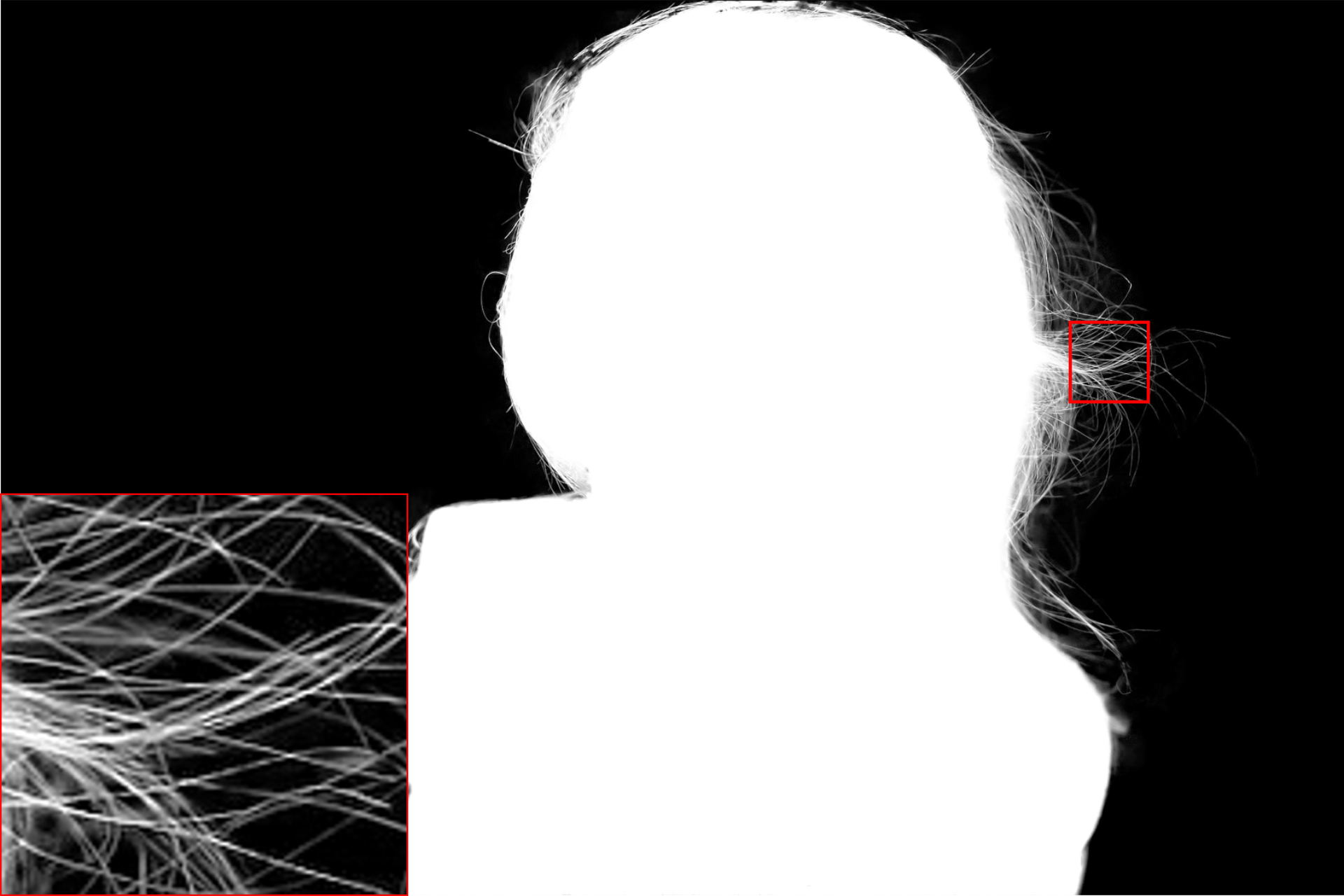} &
				\includegraphics[width=0.16\linewidth]{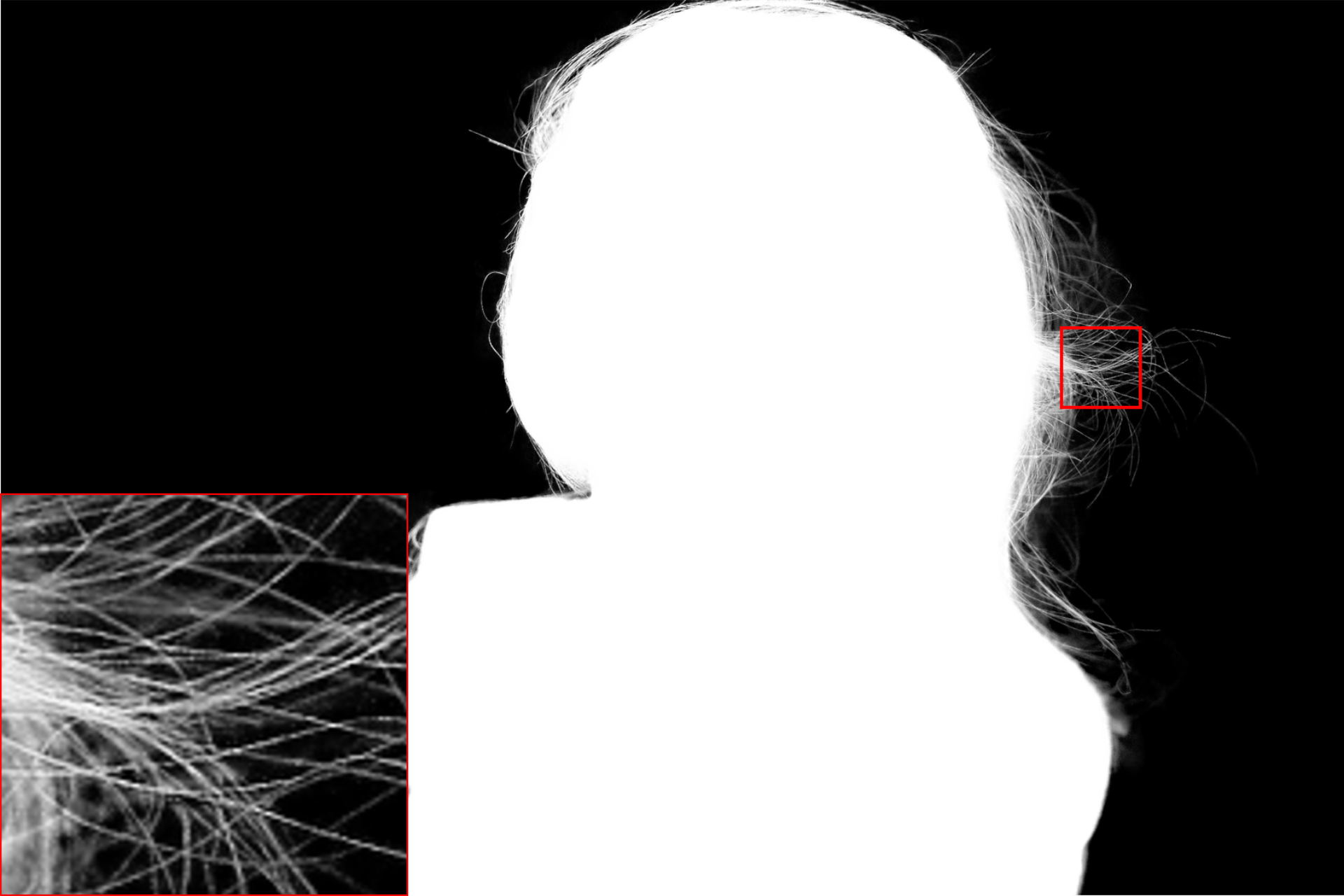} &
				\includegraphics[width=0.16\linewidth]{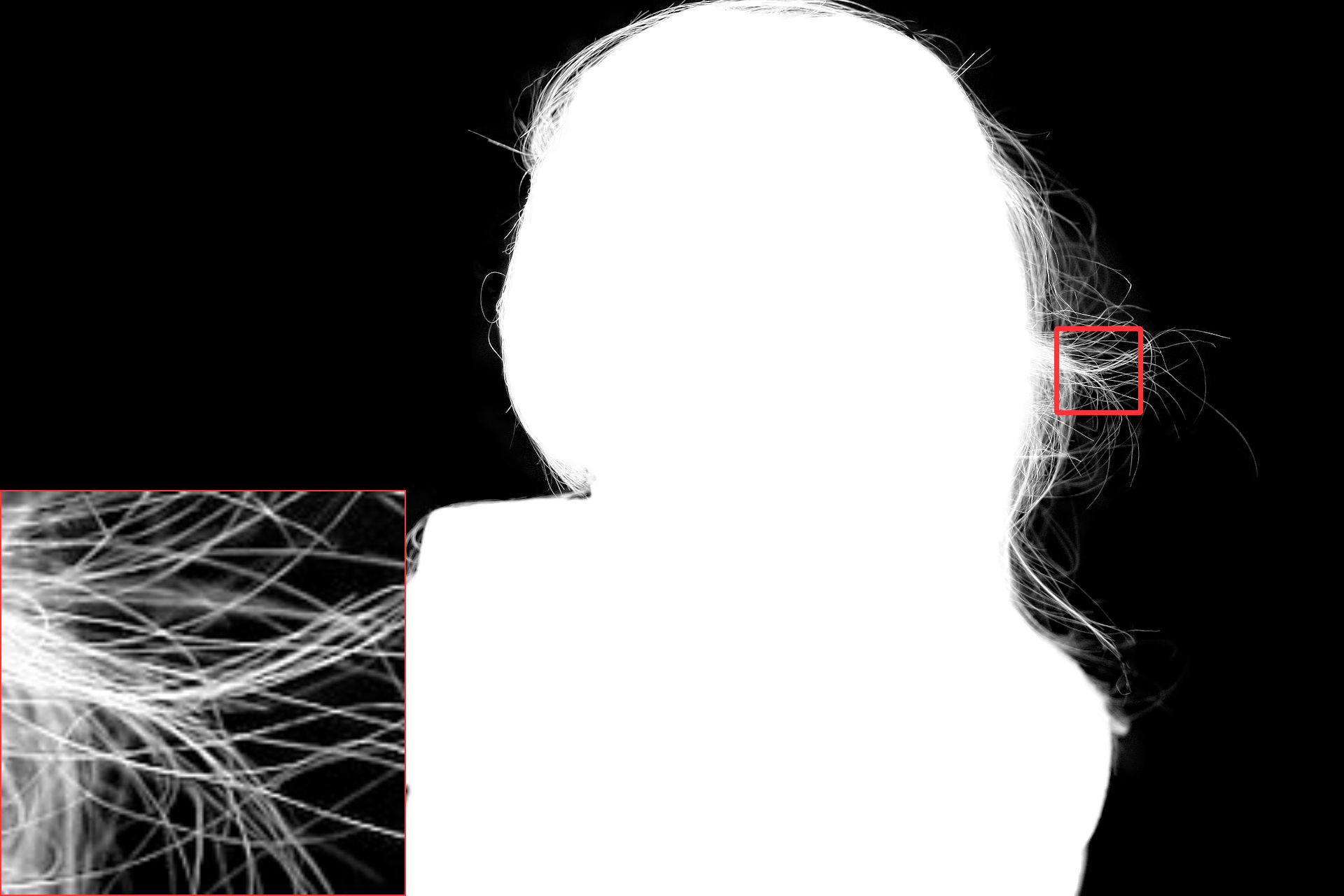} \\

				\small Input Image & Trimap & GCA~\cite{li2020natural} & A$^{2}$U~\cite{dai2021learning} & Ours & Ground Truth \\
		\end{tabular}}
	\end{center}
	\vspace{-5mm}
	\caption{The visual comparisons with GCA~\cite{li2020natural} and A$^{2}$U~\cite{dai2021learning} on the Adobe Composition-1K dataset~\cite{Xu2017Deep}. The regions marked by the red rectangles are zoomed in at the corner. }
	\vspace{-3mm}
	\label{fig:com_adobe}
\end{figure*}
\vspace{-2mm}
\subsection{Results on Adobe Composition-1k}
\label{ssec:re_com}
\vspace{-1mm}
The Adobe Composition-1k dataset is constructed by~\cite{Xu2017Deep}, which contains $431$ different foreground images and corresponding alpha mattes for training while $50$ for testing. We refer to the composition rules in~\cite{Xu2017Deep} to generate a large-scale dataset with 43100 training images and 1000 testing examples. Here we compare I2GFP with CF\cite{Levin2007A} and deep learning models, including DCNN\cite{Cho2016Natural}, DIM~\cite{Xu2017Deep}, Sample~\cite{Tang_2019_CVPR}, CA~\cite{hou2019context}, Index~\cite{hao2019indexnet}, AdaM~\cite{cai2019disentangled}, GCA~\cite{li2020natural}, A$^{2}$U~\cite{dai2021learning}, Late Fusion~\cite{Zhang2019CVPR}, and HAttMatt~\cite{Qiao_2020_CVPR}. % Xu~\et~also proposed their composition rules to generate a large-scale matting dataset with $43100$ training images and $1000$ test examples. The training and test background images and are from~\cite{Lin2014Microsoft} and~\cite{pascal-voc-2007} respectively. We compare I2GFP with some traditional methods and most existing deep-learning-based matting models on the Adobe Composition-1k dataset, including DIM~\cite{Xu2017Deep}, AlphaGAN~\cite{Shahrian2013Improving}, SampleNet~\cite{Tang_2019_CVPR}, Context-Aware~\cite{hou2019context}, IndexNet~\cite{hao2019indexnet}, AdaMatting~\cite{cai2019disentangled}, GCA~\cite{li2020natural}, A$^{2}$U~\cite{dai2021learning}, Late Fusion~\cite{Zhang2019CVPR}, and HAttMatt~\cite{Qiao_2020_CVPR}. Only Late Fusion~\cite{Zhang2019CVPR} and HAttMatt~\cite{Qiao_2020_CVPR} are trimap-free methods but they have bottlenecks in adaptability as mentioned by themselves.

\begin{table}[t]
	\centering
	\caption{The quantitative results on the Adobe Composition-1K testing dataset~\cite{Xu2017Deep}. ``Base"-baseline model, ``Base+IC"-add the IC module on the baseline. The combination of IC and GFP on the baseline is the full I2GFP model.} % All metrics are calculated inside the transition regions.
	\label{tab:quantitative_adobe}
	\begin{tabular}{l|ccccccccc}
		\hline
		Methods & CF\cite{Levin2007A} & DCNN\cite{Cho2016Natural} & & DIM\cite{Xu2017Deep} & Sample\cite{Tang_2019_CVPR} & & CA\cite{hou2019context} & Index\cite{hao2019indexnet} & AdaM\cite{cai2019disentangled} \\
		\hline
		SAD$\downarrow$ & 121.18 & 122.40 & & 50.4 & 40.35 & & 35.8 & 45.8 & 41.7 \\
		MSE$\downarrow$ & 0.076 & 0.079 & & 0.014 & 0.0099 & & 0.0082 & 0.013 & 0.010 \\
		Grad$\downarrow$ & 130.63 & 129.57 & & 31.0 & - & & 17.3 & 25.9 & 16.8 \\
		Conn$\downarrow$ & 120.16 & 121.80 & & 50.8 & - & & 33.2 & 43.7 & - \\
		\hline
		& GCA\cite{li2020natural} & A$^2$U\cite{dai2021learning} & \vline & LF\cite{Zhang2019CVPR} & HAtt\cite{Qiao_2020_CVPR} & \vline & Base & Base+IC & I2GFP \\
		\hline
		SAD$\downarrow$ & 35.28 & \textbf{32.15} & \vline & 49.02 & 45.04 & \vline & 41.82 & 35.46 & 32.20 \\
		MSE$\downarrow$ & 0.0091 & 0.0082 & \vline & 0.020 & 0.015 & \vline & 0.0106 & 0.0084 & \textbf{0.0079} \\
		Grad$\downarrow$ & 16.92 & 16.39 & \vline & 34.33 & 21.70 & \vline & 20.76 & 17.65 & \textbf{15.64} \\
		Conn$\downarrow$ & 32.53 & \textbf{29.25} & \vline& 50.60 & 44.28 & \vline & 36.71 & 34.78 & 30.41 \\
		\hline
	\end{tabular}
\vspace{-3mm}
\end{table}
The quantitative summary is reported in Table~\ref{tab:quantitative_adobe}, and the best results are shown in bold. We can observe that I2GFP has clear advantages on the MSE and Gradient metrics compared to existing methods, corresponding to $3.66\%$ and $4.58\%$ performance promotion, respectively. These improvements are benefiting from IC module and GFP branch. IC can consolidate the opacity consistency between different foreground parts via multi-scale feature communication. GFP can provide rich appearances to complement details and texture to further visual quality. By contrast, on the SAD and Conn metrics, I2GFP is slightly inferior to A$^{2}$U matting~\cite{dai2021learning}. The potential analysis is that the affinity-aware functions in~\cite{li2020natural} can retain many sampling details. It is noted that most deep learning models~\cite{Tang_2019_CVPR,hou2019context,hao2019indexnet,cai2019disentangled,li2020natural,Zhang2019CVPR,Qiao_2020_CVPR} employ sophisticated backbones to extract more advanced semantics, and the performance of I2GFP indicates that wider and higher feature representation contributes better to alpha perception.

As shown in Fig.~\ref{fig:com_adobe}, we make the visual comparisons with GCA~\cite{li2020natural} and A$^{2}$U~\cite{dai2021learning} matting. The I2GFP demonstrates more details and textures for common transition regions, like hairs and nets. The better visual presentation is mainly due to the GFP branch, which can supply more exquisite appearances. The high-resolution encoder-decoder motivation can also retain more primitive image features. Our VGG-16 backbone has limited ability to capture semantics as verified by many classification tasks and the improvement is basically from IC and GFP. Thus we can conclude that advanced semantics may not be the consequential factor for image matting.

\begin{figure}[t]
	\begin{center}
		\setlength{\tabcolsep}{0.4pt}{
			\begin{tabular}{ccccc}
				\includegraphics[width=0.192\linewidth]{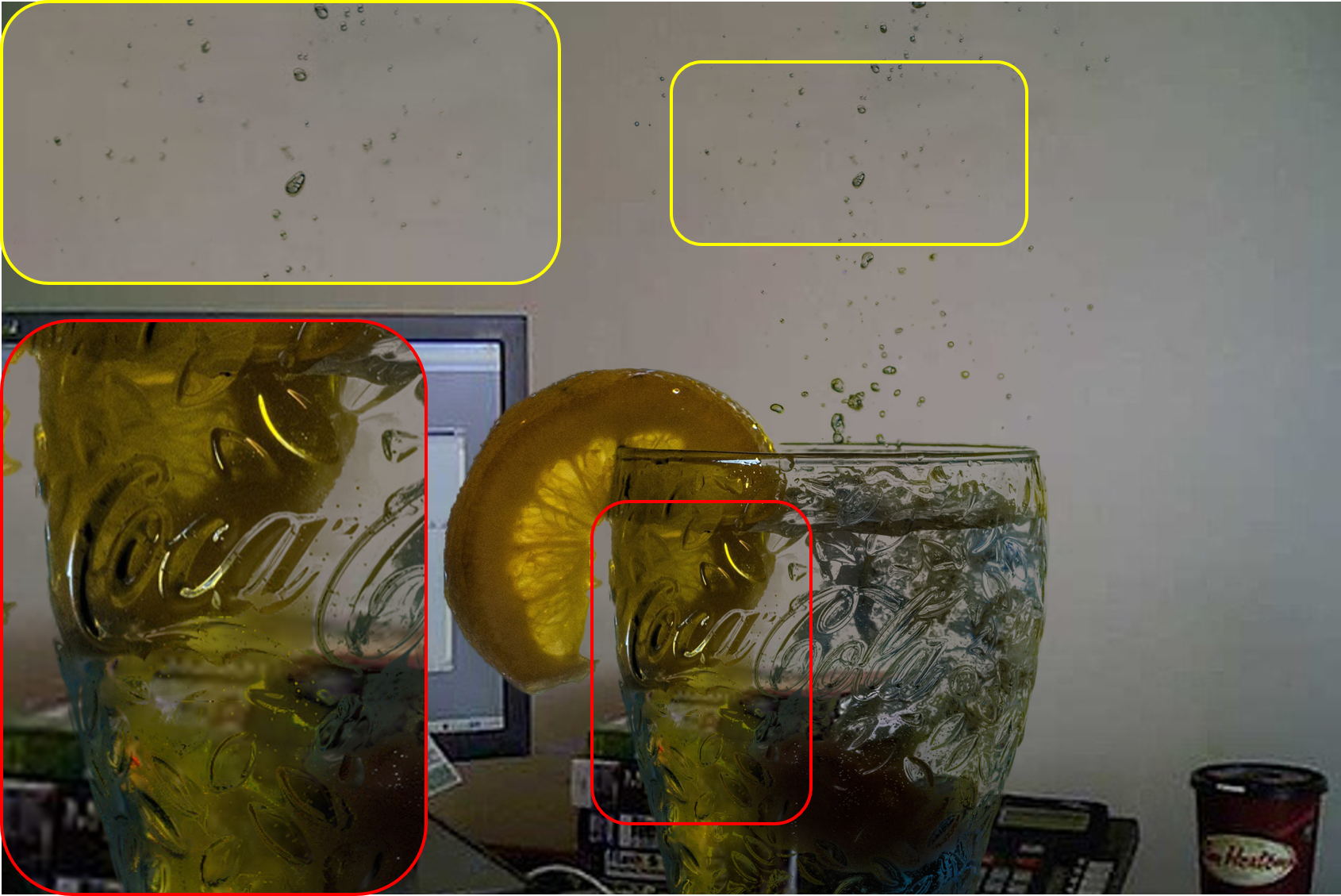} &
				\includegraphics[width=0.192\linewidth]{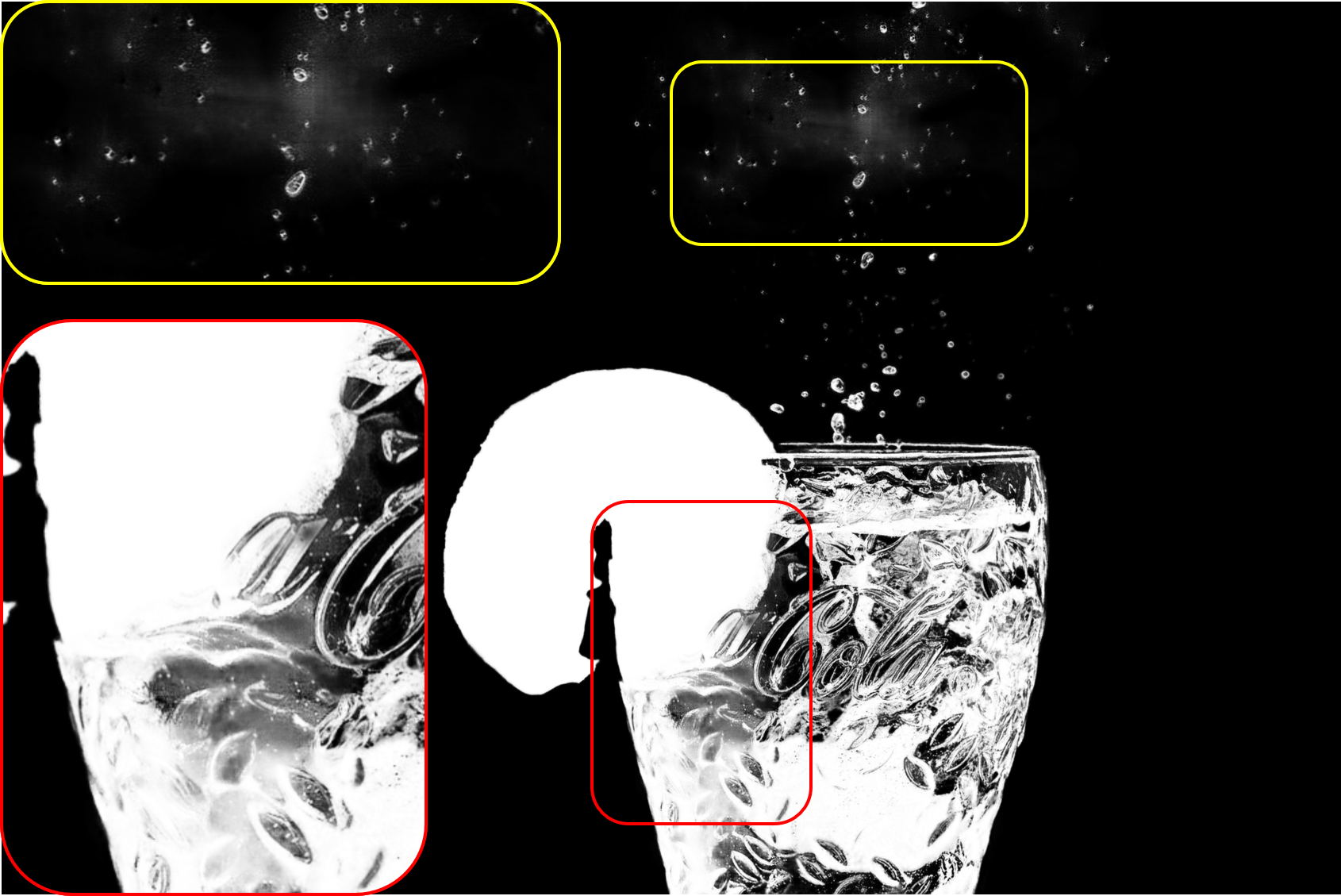} &
				\includegraphics[width=0.192\linewidth]{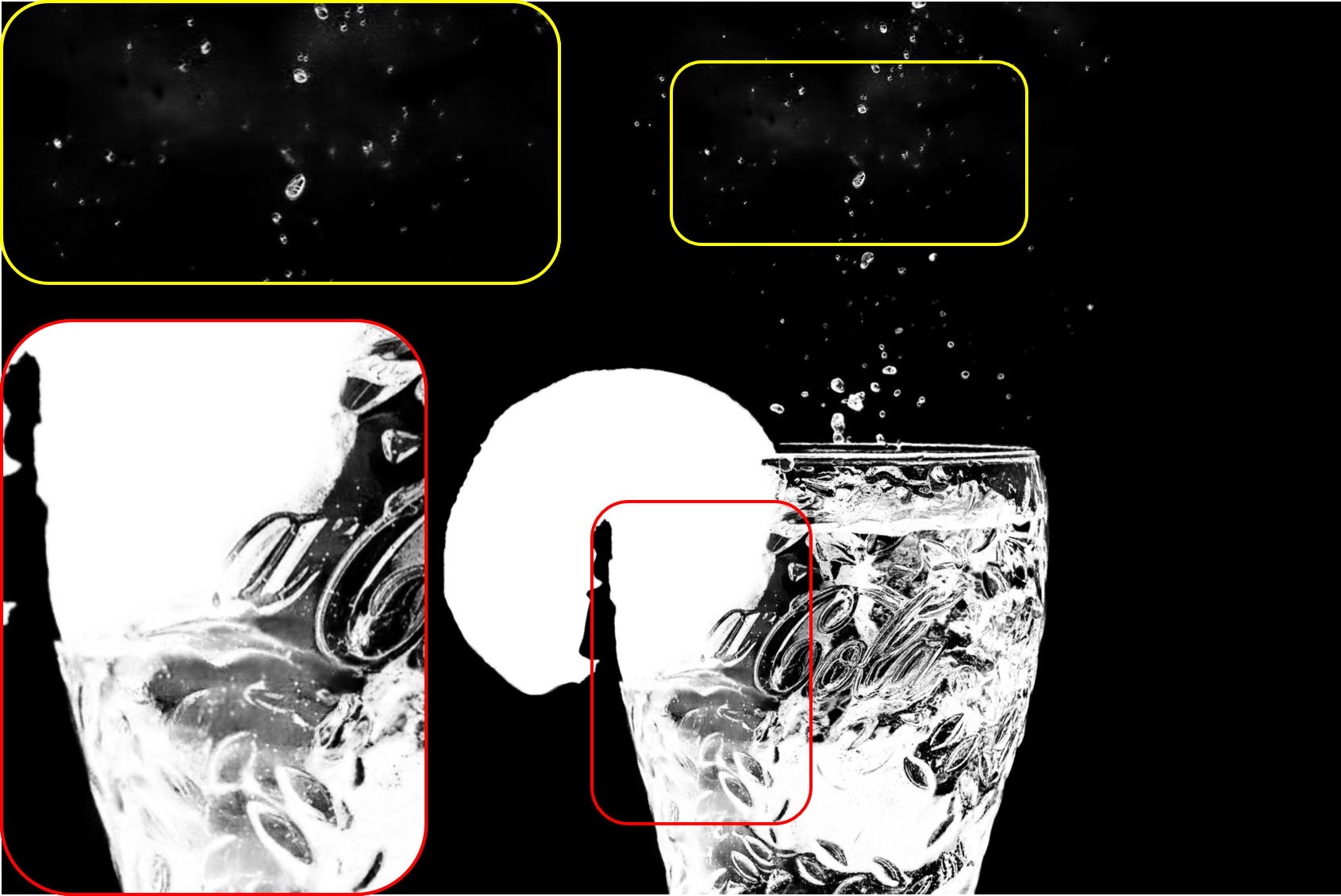} &
				\includegraphics[width=0.192\linewidth]{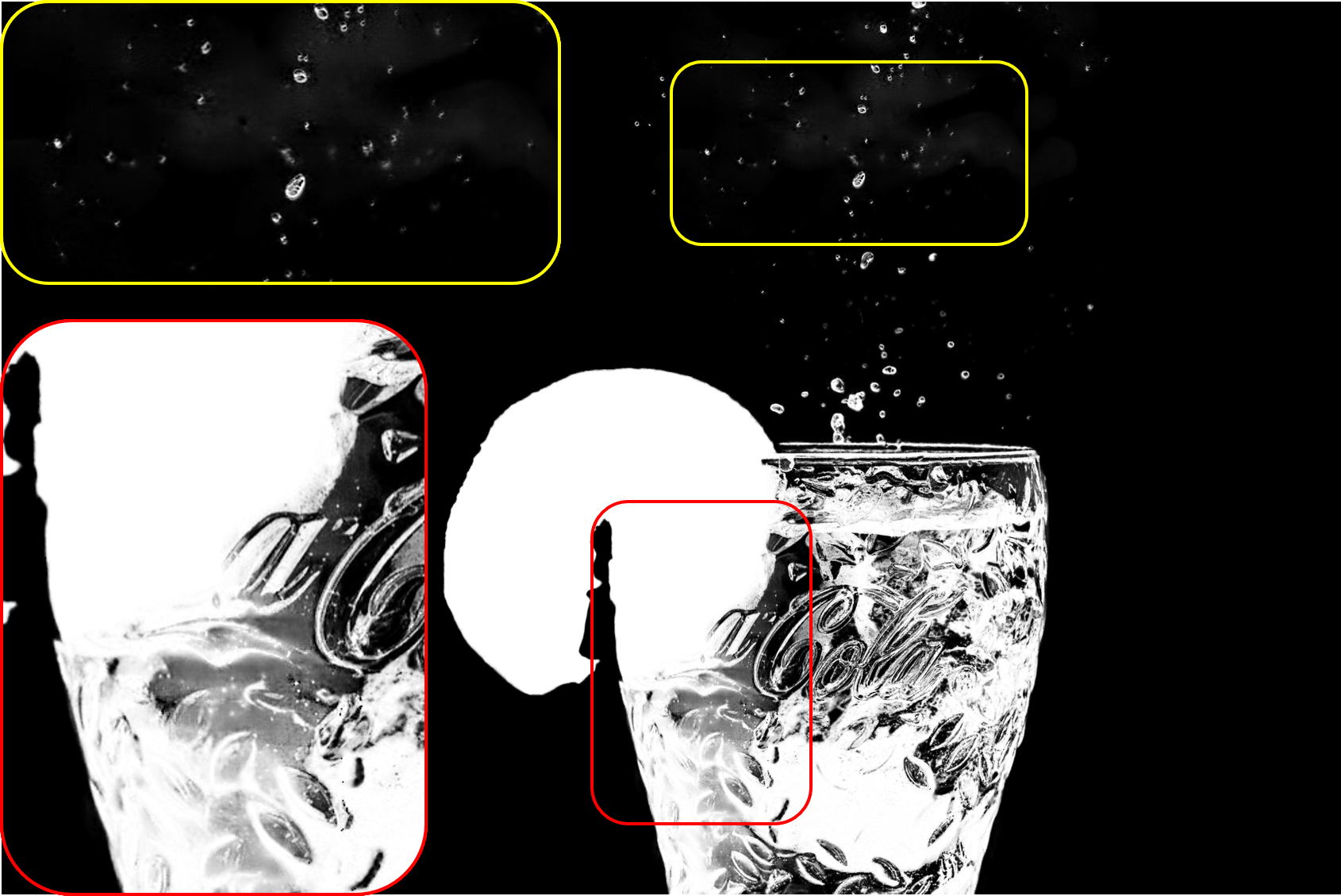} &
				\includegraphics[width=0.192\linewidth]{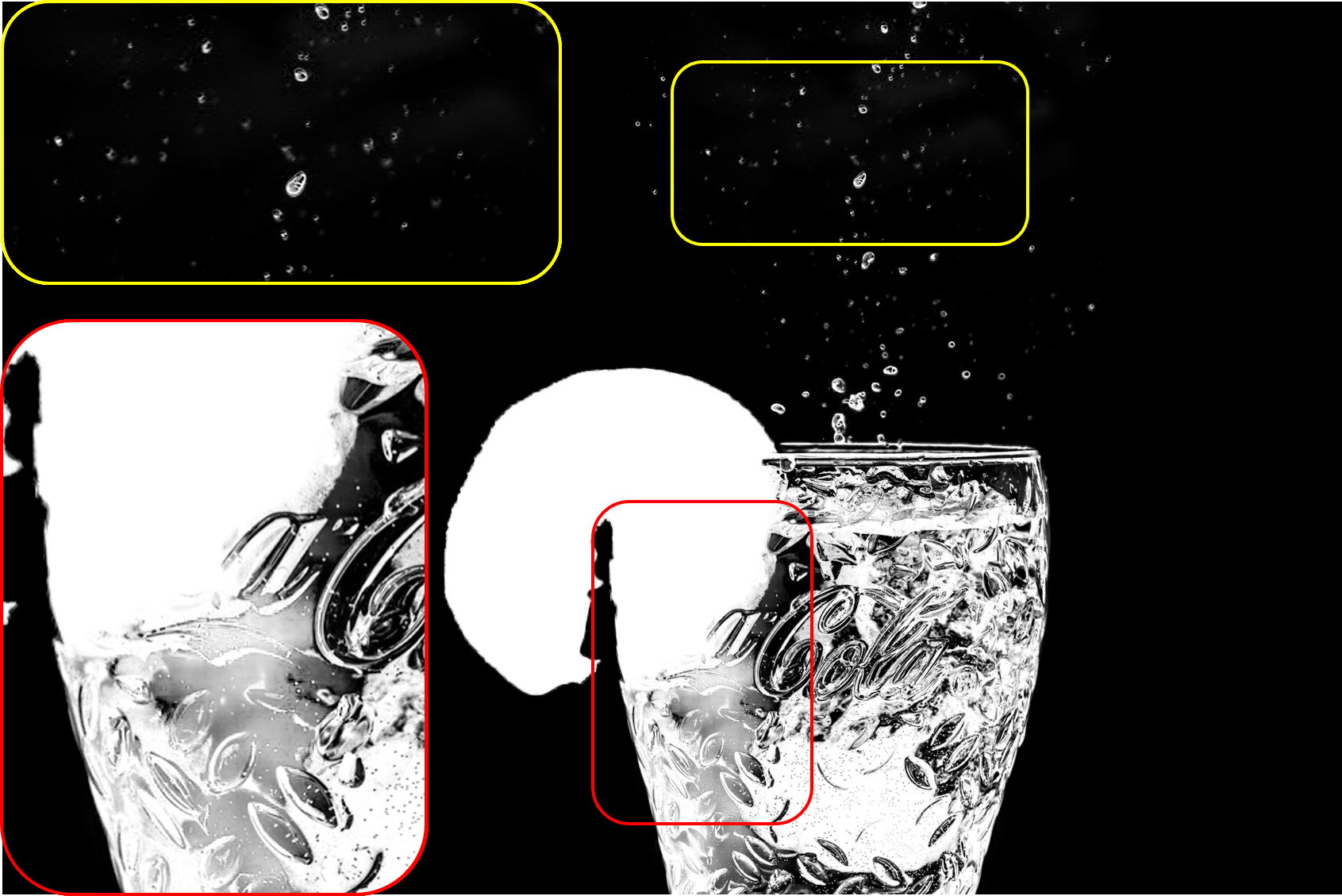} \\
				
				Input Image & Base & Base+IC & I2GFP & Ground Truth \\
		\end{tabular}}
	\end{center}
	\vspace{-5mm}
	\caption{Ablation study: the suspended water droplets and the lines on the glass gradually clarify with the model components incremented (zoomed regions on the left). }
	\vspace{-1mm}
	\label{fig:ablation}
\end{figure}

\vspace{-2mm}
\subsection{Ablation Study}
\label{ssec:ablation}
\vspace{-1mm}
Here we perform an ablation study to verify different components. All models here are trained and evaluated on the Adobe dataset~\cite{Xu2017Deep}. The $output\_stride = 4$ encoder is cascaded with a U-Net like decoder to form the baseline model. As reported in Tab.~\ref{tab:quantitative_adobe}, the baseline model can achieve better results than DIM~\cite{Xu2017Deep} and Index~\cite{hao2019indexnet}, proving the validity of the higher motivation ($output\_stride = 4$ can retain more primitive attributes). Then we add IC on the decoder stage to promote communications between different-level encoder features. IC can provide wider feature fields for each layer, and there is a noticeable improvement in all metrics after adding IC. The I2GFP is the full model with IC and GFP, which achieves the best results in Tab.~\ref{tab:quantitative_adobe}.

The visual comparison from the Adobe Composition-1k testing set is shown in Fig.~\ref{fig:ablation}. Both IC and GFP can supply more details and texture, and we can observe that the suspended water droplets and the lines on the glass gradually become clear after adding different components.
\begin{table}[t]
	\caption{The quantitative results on the Distinctions-646 dataset~\cite{Qiao_2020_CVPR}. The metrics about HAtt~\cite{Qiao_2020_CVPR} are summarized on the full images. ``I2GFP-f" represents our results on the full images. The other metrics are calculated inside the transition regions. }
	\label{tab:quantitative_dis}
	\setlength{\tabcolsep}{0.5mm
	\begin{tabular}{l|ccccccccccc}
		\hline
		Methods & CF\cite{Levin2007A} & DIM\cite{Xu2017Deep} & Index\cite{hao2019indexnet} & GCA\cite{li2020natural} & \vline & HAtt\cite{Qiao_2020_CVPR} & \vline & Base & Base+IC & I2GFP & I2GFP-f \\
		\hline
		SAD$\downarrow$ & 146.66 & 44.15 & 34.47 & 26.59 & \vline & 48.98 & \vline & 31.12 & 28.99 & \textbf{26.48} & 30.57 \\
		MSE$\downarrow$ & 0.141 & 0.031 & 0.019 & 0.015 & \vline & 0.009 & \vline & 0.015 & 0.013 & \textbf{0.012} & 0.005 \\
		Grad$\downarrow$ & 356.50 & 39.08 & 28.31 & 19.50 & \vline & 41.57 & \vline & 18.86 & 15.40 & \textbf{15.19} & 21.45 \\ 
		Conn$\downarrow$ & 143.12 & 44.65 & 33.37 & \textbf{25.23} & \vline & 49.93 & \vline & 31.62 & 28.52 & 26.39 & 30.91 \\
		\hline
	\end{tabular}}
\end{table}
\begin{figure*}[t]
	\begin{center}
		\setlength{\tabcolsep}{0.6pt}{
			\begin{tabular}{cccccc}			
				\includegraphics[width=0.16\linewidth]{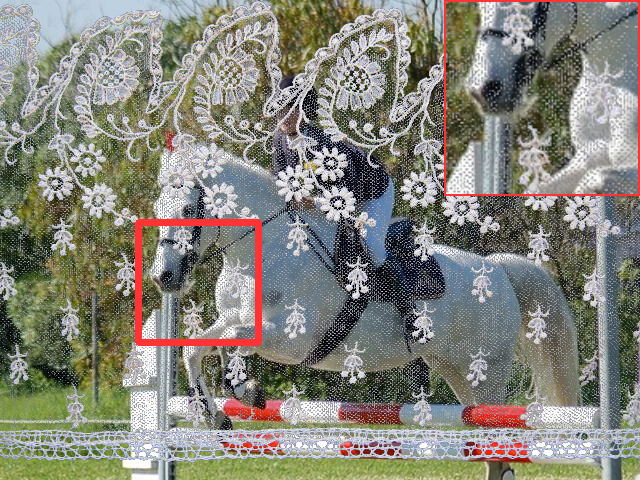} &
				\includegraphics[width=0.16\linewidth]{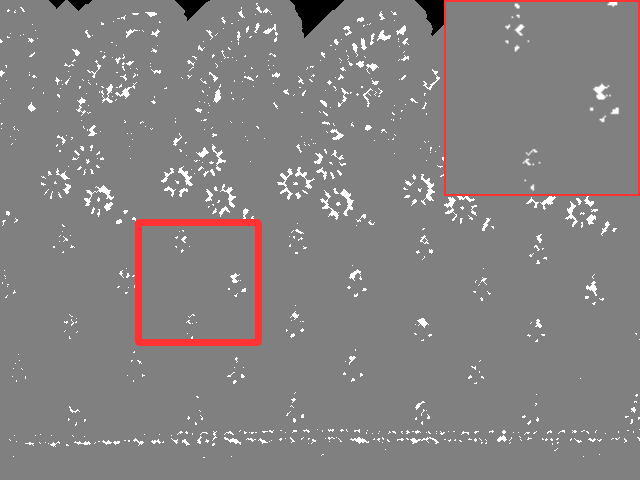} &
				\includegraphics[width=0.16\linewidth]{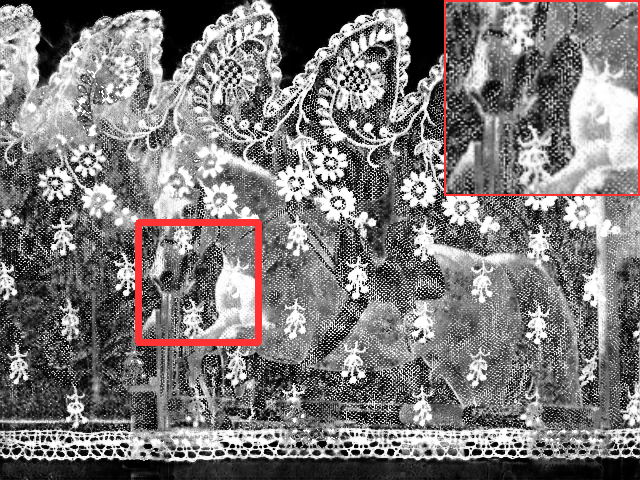} &
				\includegraphics[width=0.16\linewidth]{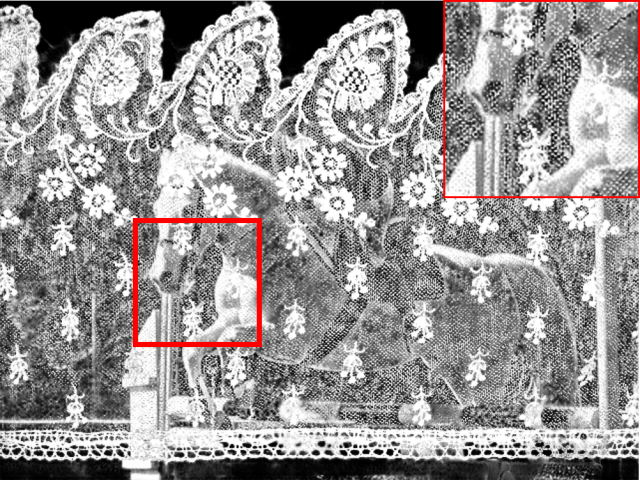} &
				\includegraphics[width=0.16\linewidth]{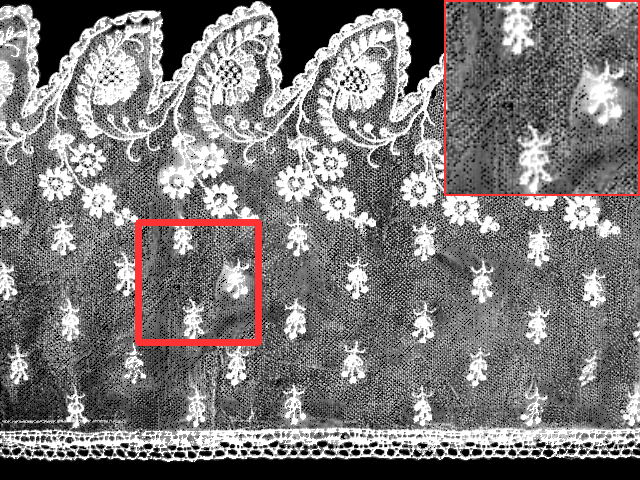} &
				\includegraphics[width=0.16\linewidth]{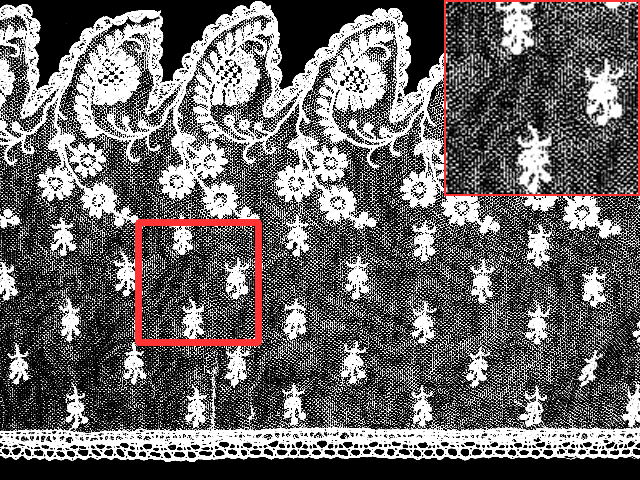} \\
				
				\includegraphics[width=0.16\linewidth]{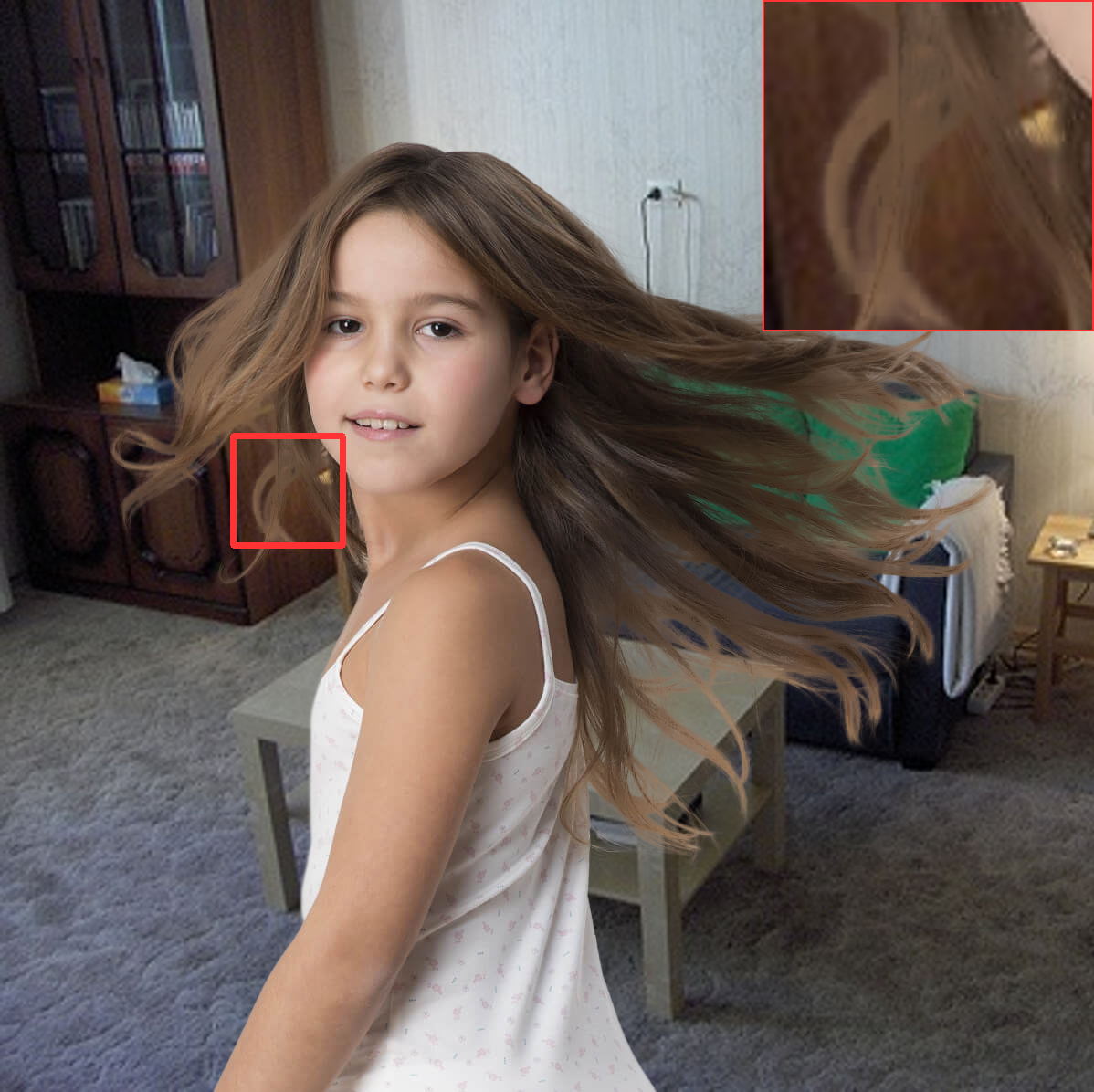} &
				\includegraphics[width=0.16\linewidth]{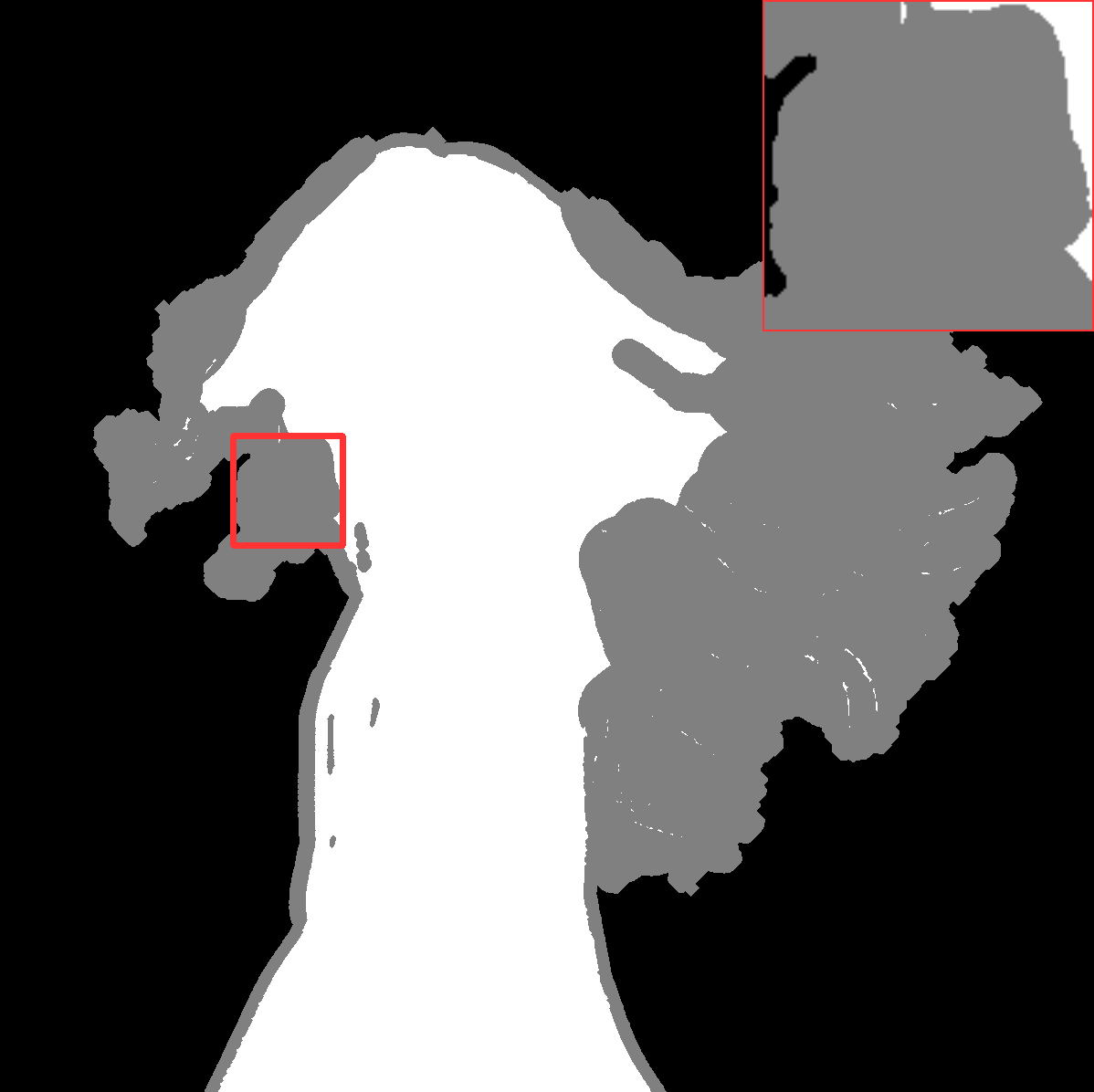} &
				\includegraphics[width=0.16\linewidth]{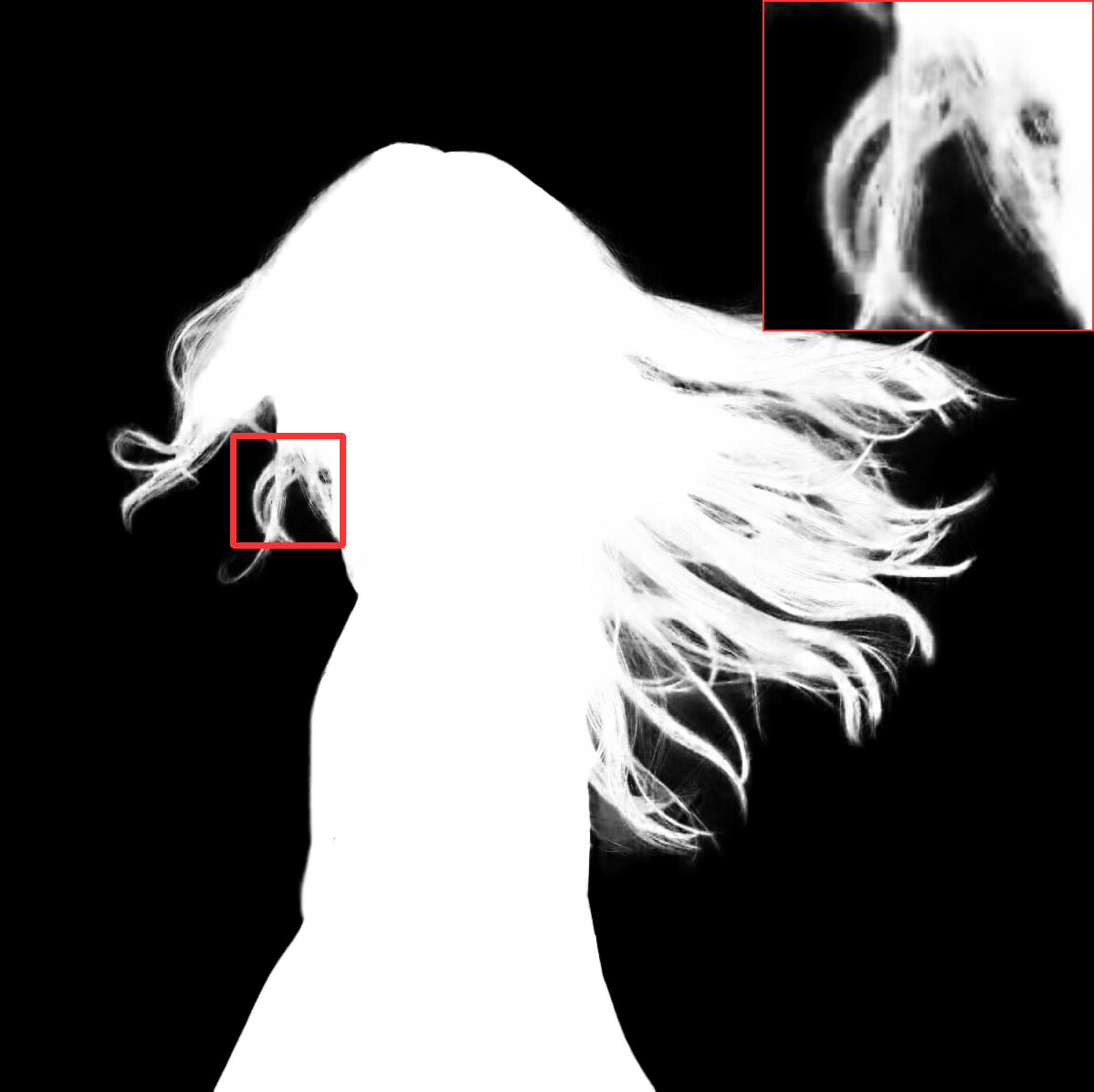} &
				\includegraphics[width=0.16\linewidth]{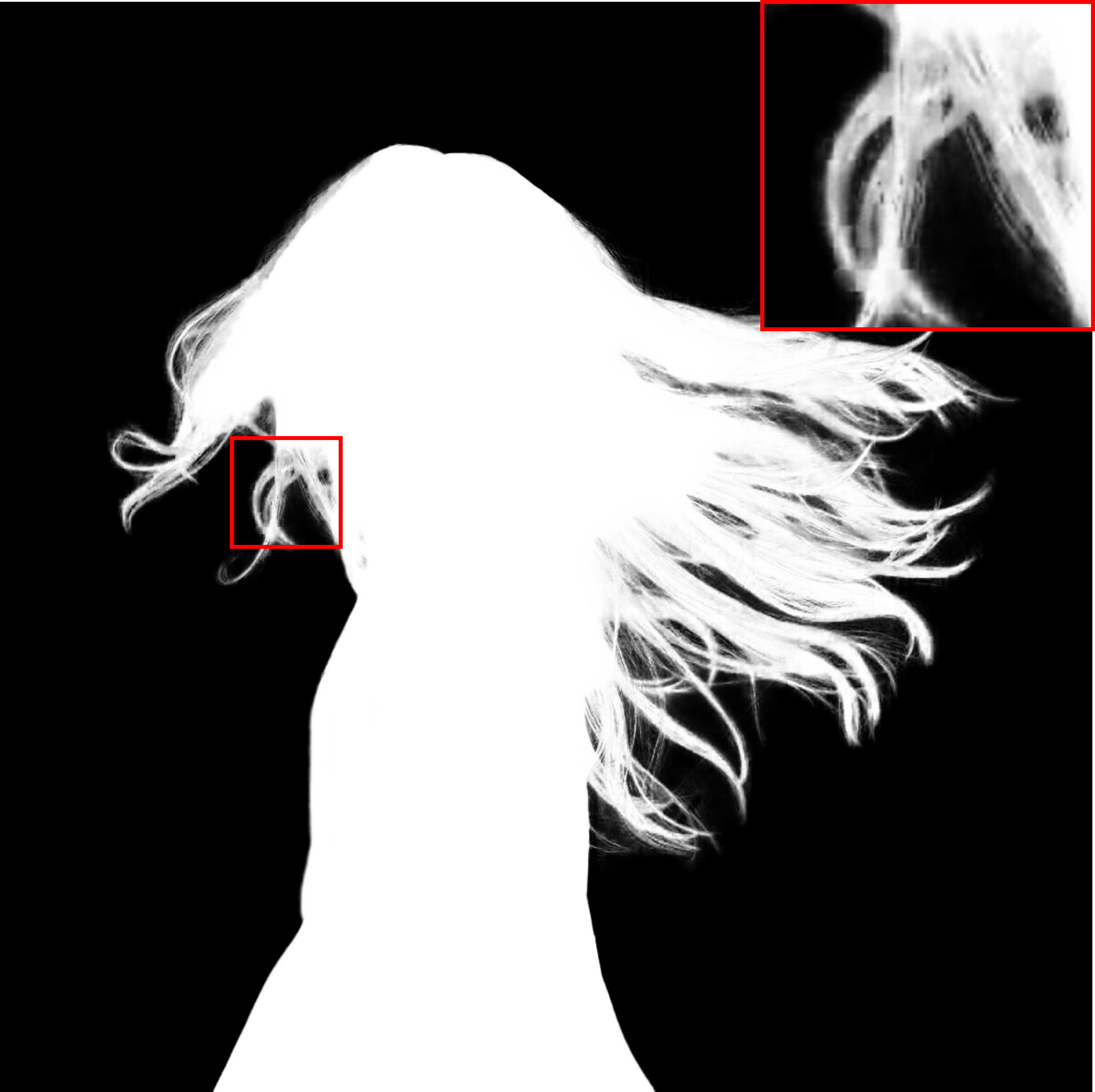} &
				\includegraphics[width=0.16\linewidth]{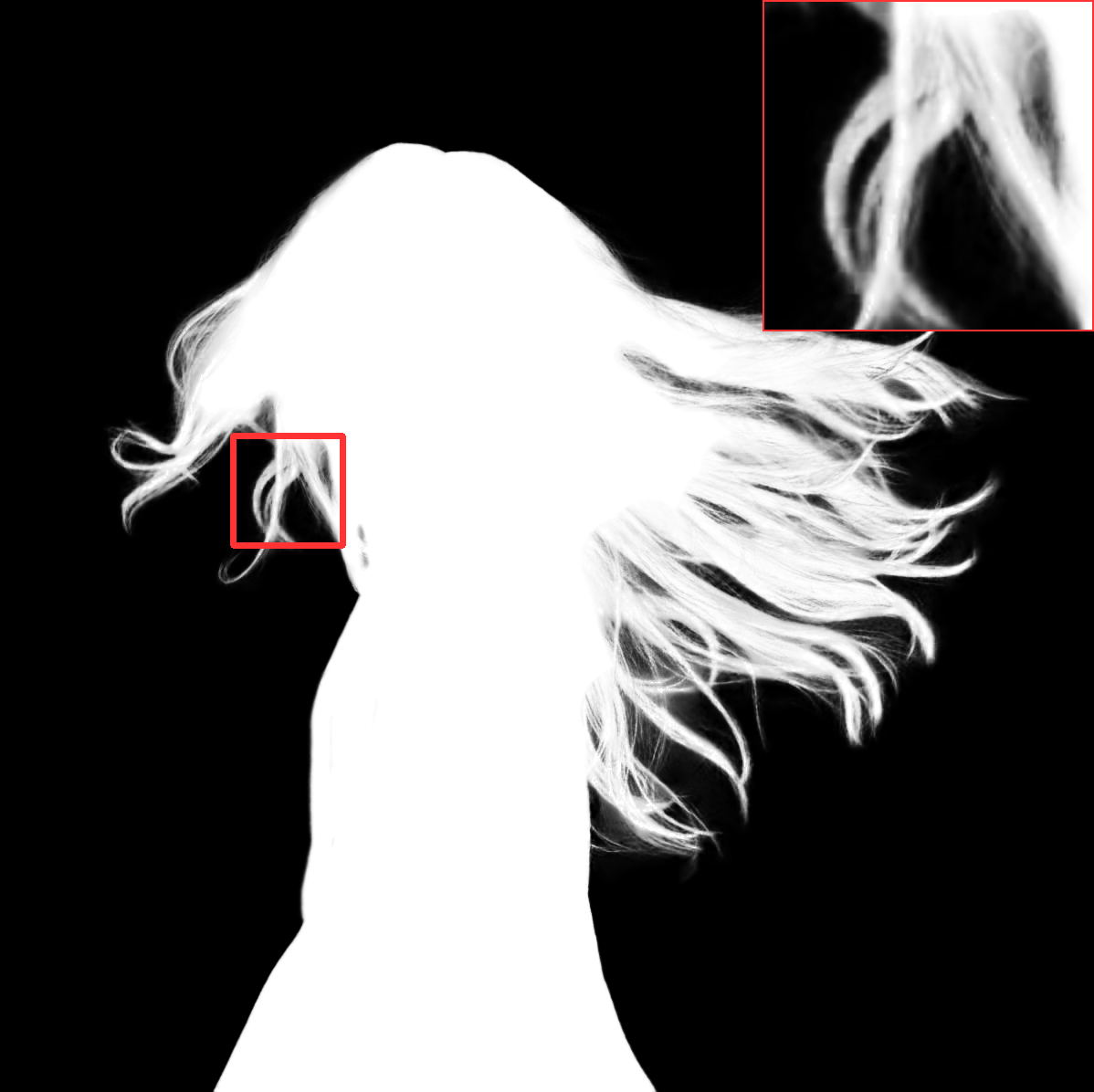} &
				\includegraphics[width=0.16\linewidth]{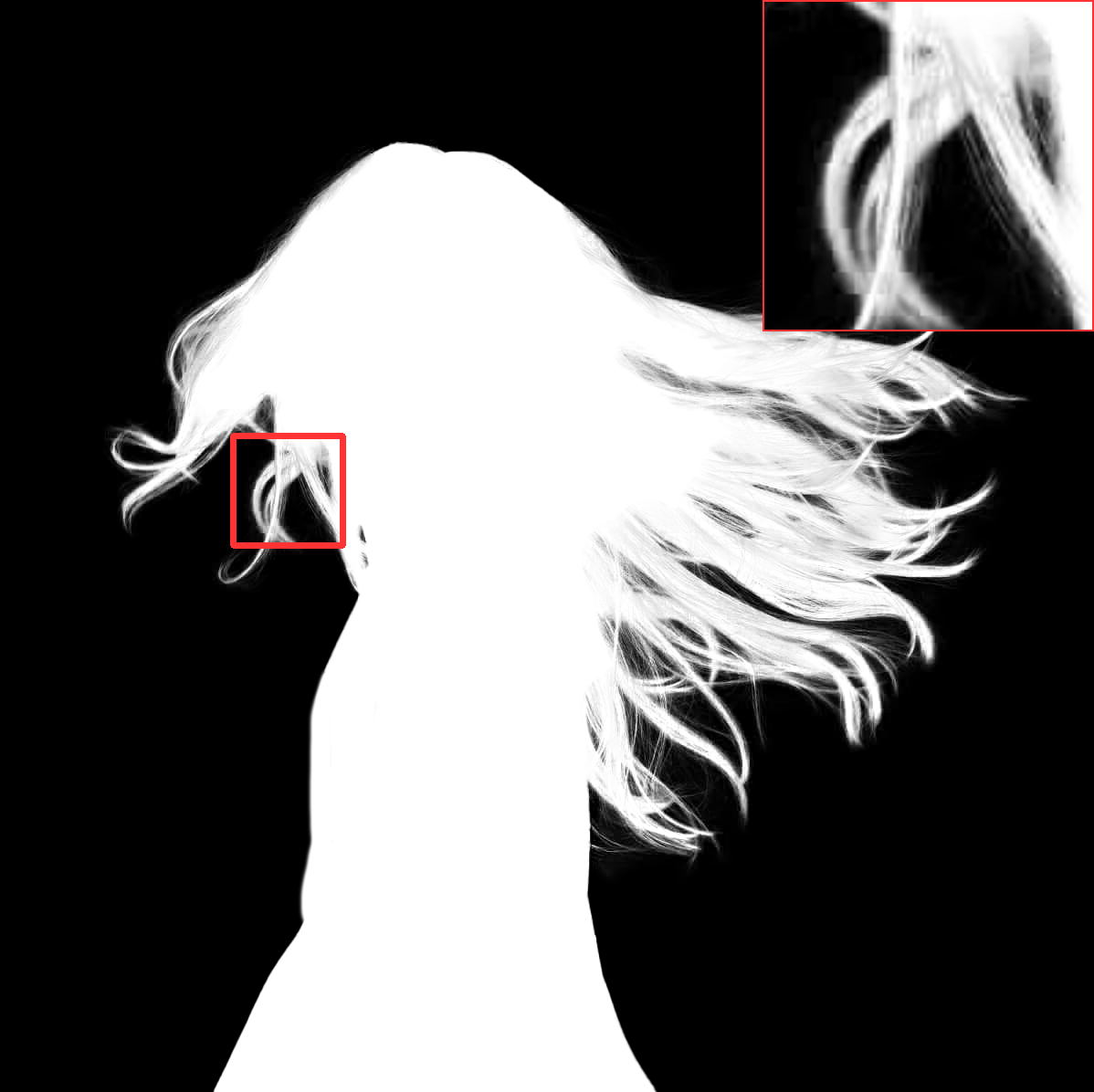} \\

				\small Input Image & Trimap & Index~\cite{hao2019indexnet} & GCA~\cite{li2020natural} & Ours & Ground Truth \\
		\end{tabular}}
	\end{center}
	\vspace{-5mm}
	\caption{The visual comparisons with IndexNet~\cite{hao2019indexnet} and GCA~\cite{li2020natural} on the Distinctions-646 dataset~\cite{Qiao_2020_CVPR}. The regions marked by the red rectangles are zoomed in at the corner of the alpha mattes to show more details. }
	\vspace{-3mm}
	\label{fig:com_dis}
\end{figure*}

\vspace{-2mm}
\subsection{Results on Distinctions-646}
\label{ssec:re_dis}
\vspace{-1mm}
The Distinctions-646 dataset~\cite{Qiao_2020_CVPR} consists of $596$ distinct training alpha mattes and $50$ testing ones. We compare I2GFP with some available methods, CF~\cite{Levin2007A}, DIM~\cite{Xu2017Deep}, Index~\cite{hao2019indexnet}, GCA~\cite{li2020natural}, and HAtt~\cite{Qiao_2020_CVPR}. The quantitative and qualitative results are shown in Tab.~\ref{tab:quantitative_dis} and Fig.~\ref{fig:com_dis}.

Compared to GCA, I2GFP improves $20\%$ on MSE, $22.1\%$ on Gradient, and is slightly worse than GCA on Connectivity. We can conclude that the wider and higher feature representation is robust and generalized for different datasets. The visual demonstration in Fig.~\ref{fig:com_dis} can clearly indicate the deficiency of Index and GCA. They both show poor performance on the net-like foreground, as shown in the first row of Fig.~\ref{fig:com_dis}. Many holes in the net are fuzzed together, making it difficult to distinguish them directly even are zoomed in. With the detailed appearances provided by GFP, our model can handle such a situation impressively, and the textures of the nets are visible.
\begin{figure*}[t]
	\begin{center}
		\begin{tabular}{c}
			\includegraphics[width=0.98\linewidth]{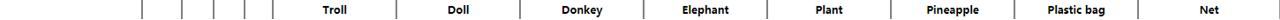} \\
			...... \\
			\includegraphics[width=0.98\linewidth]{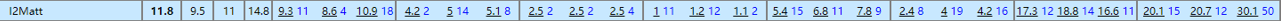} \\
		\end{tabular}
	\end{center}
	\vspace{-5mm}
	\caption{A screenshot from the benchmark website~\cite{rhemann2009perceptually}. I2GFP gets an overall score of 11.8 and rank seventh among all published papers. Only our method employs a simple backbone and produces better results through wider and higher features. }
	\vspace{-3mm}
	\label{fig:bench_rank}
\end{figure*}

\begin{figure*}[t]
	\begin{center}
		\setlength{\tabcolsep}{0.6pt}{
			\begin{tabular}{cccccc}
				\includegraphics[height=0.116\linewidth]{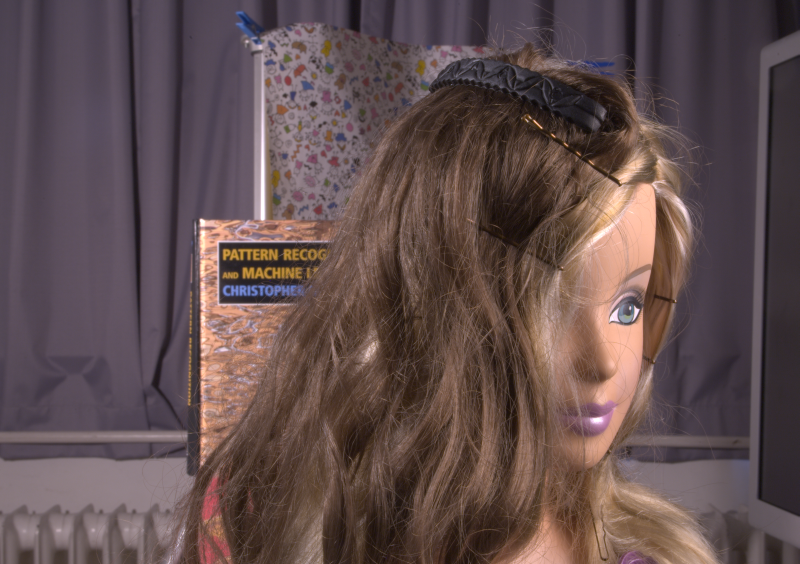} &
				\includegraphics[height=0.116\linewidth]{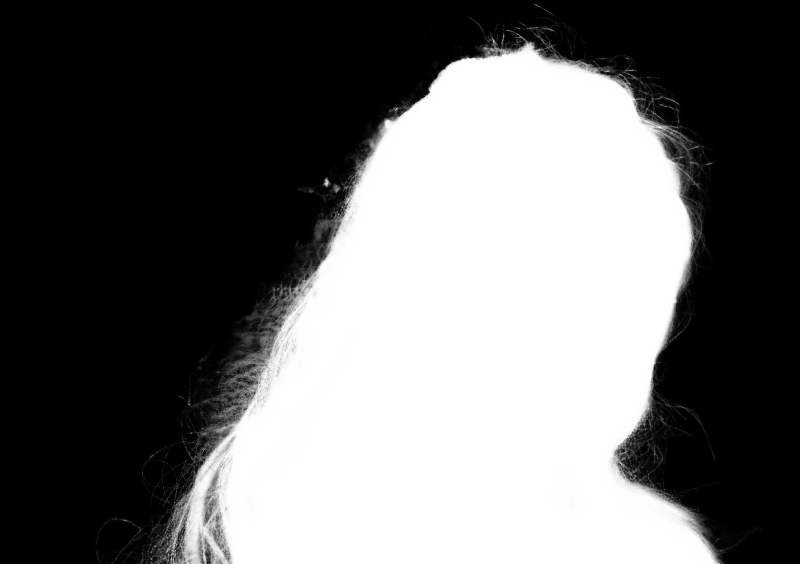} &
				\includegraphics[height=0.116\linewidth]{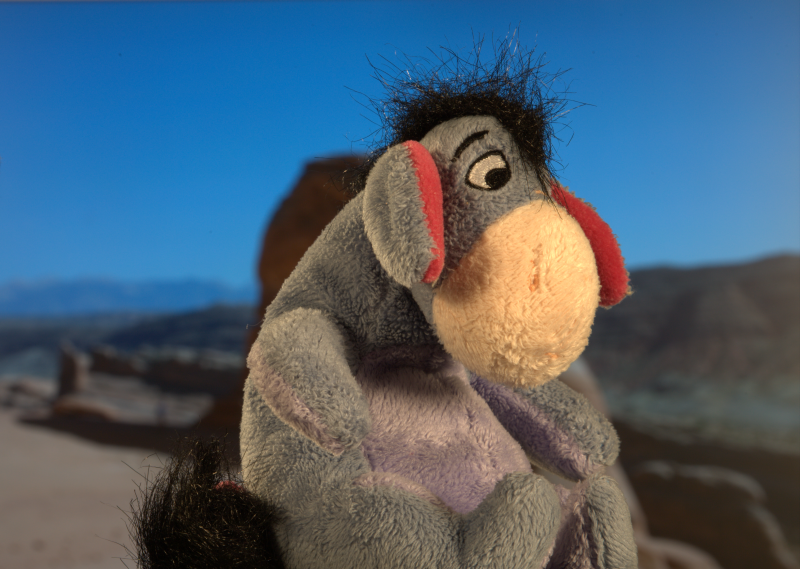} &
				\includegraphics[height=0.116\linewidth]{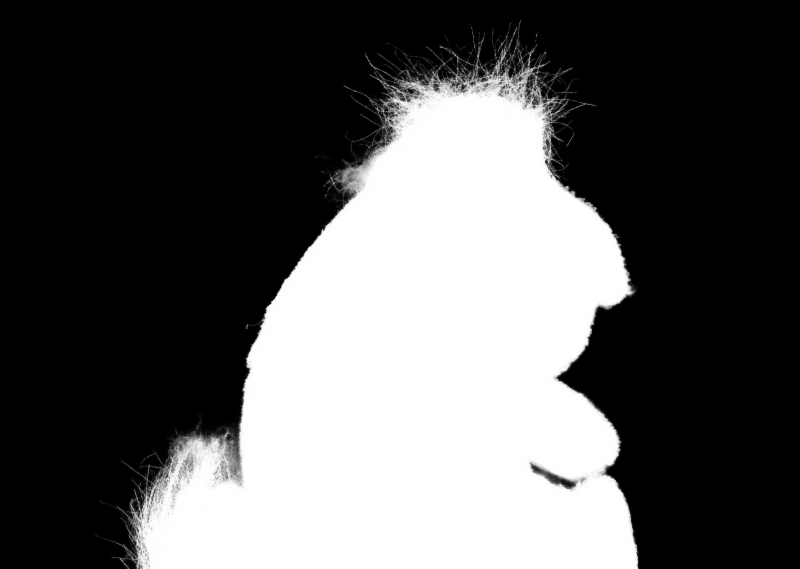} &
				
				\includegraphics[height=0.116\linewidth]{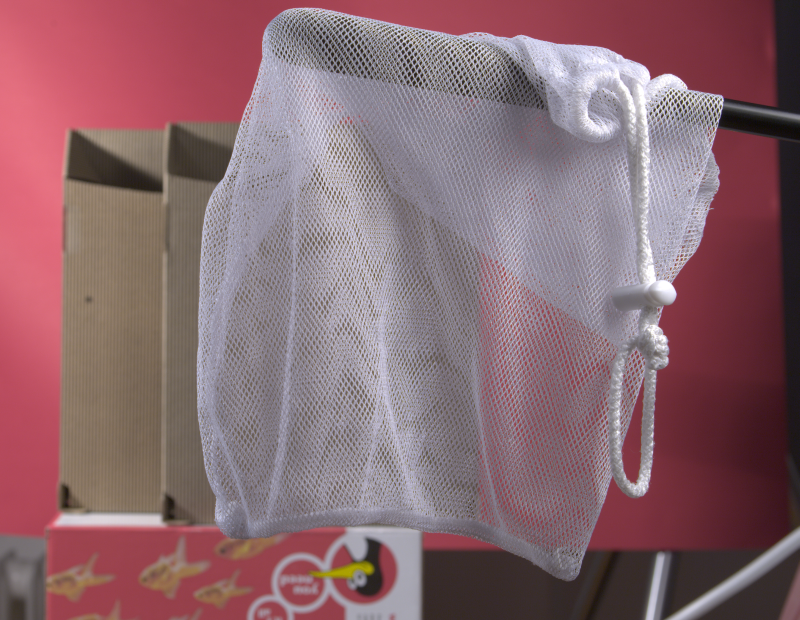} &
				\includegraphics[height=0.116\linewidth]{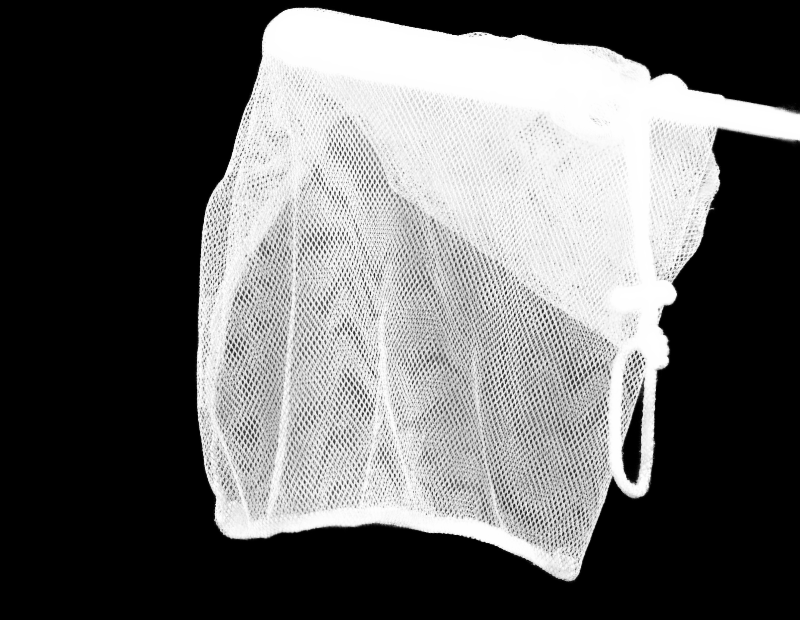} \\
				%\includegraphics[width=0.16\linewidth]{./figures/alpha_bench/elephant.png} &
				%\includegraphics[width=0.16\linewidth]{./figures/alpha_bench/tri_elephant.png} &
				%\includegraphics[width=0.16\linewidth]{./figures/alpha_bench/p_elephant.png} \\
				
%				\includegraphics[width=0.16\linewidth]{./figures/alpha_bench/plant.png} &
%				\includegraphics[width=0.16\linewidth]{./figures/alpha_bench/tri_plant.png} &
%				\includegraphics[width=0.16\linewidth]{./figures/alpha_bench/p_plant.png} &
%				\includegraphics[width=0.16\linewidth]{./figures/alpha_bench/pineapple.png} &
%				\includegraphics[width=0.16\linewidth]{./figures/alpha_bench/tri_pineapple.png} &
%				\includegraphics[width=0.16\linewidth]{./figures/alpha_bench/p_pineapple.png} \\
%				
%				\includegraphics[width=0.16\linewidth]{./figures/alpha_bench/troll.png} &
%				\includegraphics[width=0.16\linewidth]{./figures/alpha_bench/tri_troll.png} &
%				\includegraphics[width=0.16\linewidth]{./figures/alpha_bench/p_troll.png} &
%				\includegraphics[width=0.16\linewidth]{./figures/alpha_bench/plasticbag.png} &
%				\includegraphics[width=0.16\linewidth]{./figures/alpha_bench/tri_plasticbag.png} &
%				\includegraphics[width=0.16\linewidth]{./figures/alpha_bench/p_plasticbag.png} \\
				
				%Input Image & Trimap & Alpha Matte & Input Image & Trimap & Alpha Matte \\
		\end{tabular}}
	\end{center}
	\vspace{-5mm}
	\caption{The alpha mattes of I2GFP on the benchmark images~\cite{rhemann2009perceptually}.}
	\vspace{-3mm}
	\label{fig:alpha}
\end{figure*}

\begin{figure}[t]
	\begin{center}
		\setlength{\tabcolsep}{0.4pt}{
			\begin{tabular}{cccccc}
				\includegraphics[height=0.108\linewidth]{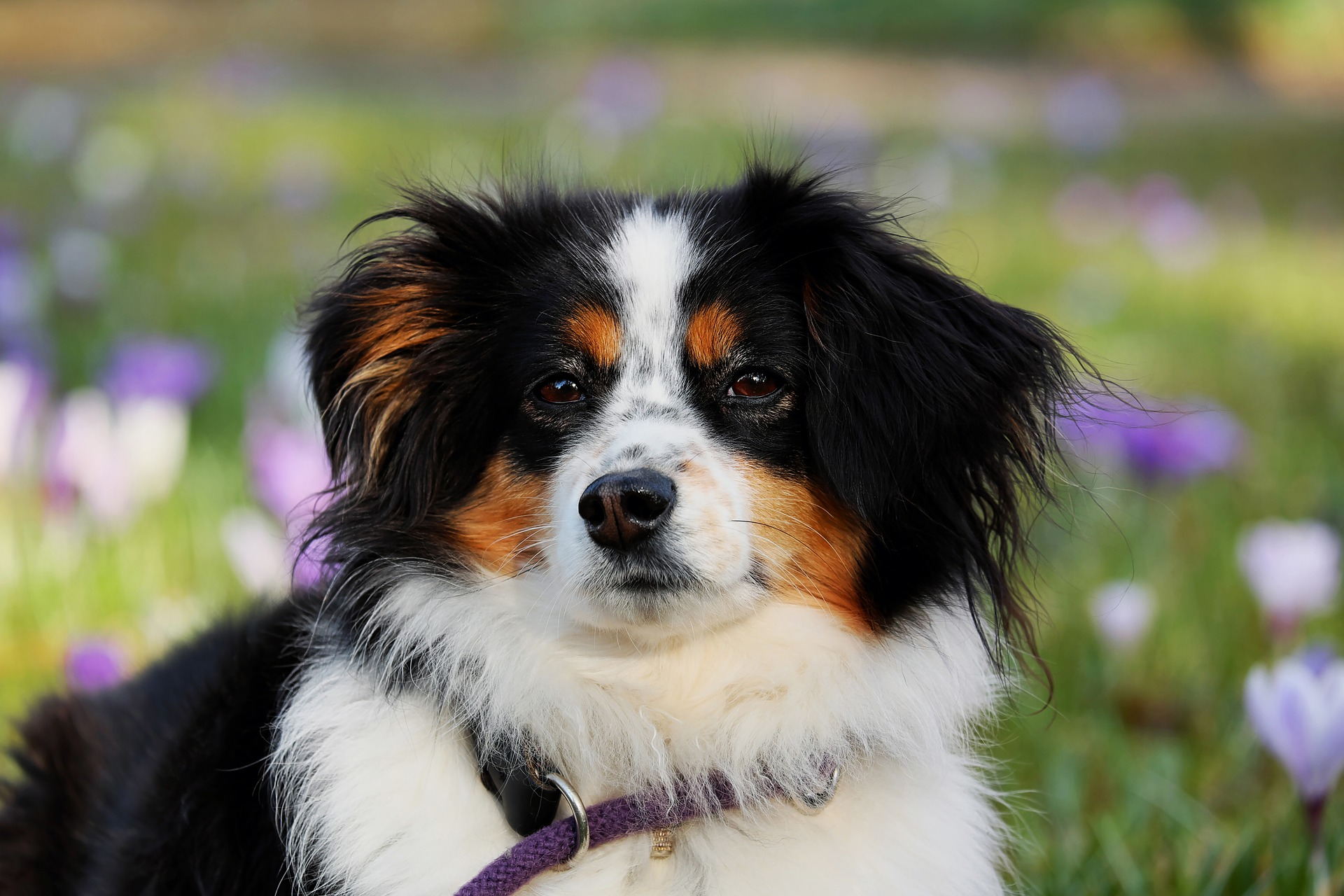} &
				\includegraphics[height=0.108\linewidth]{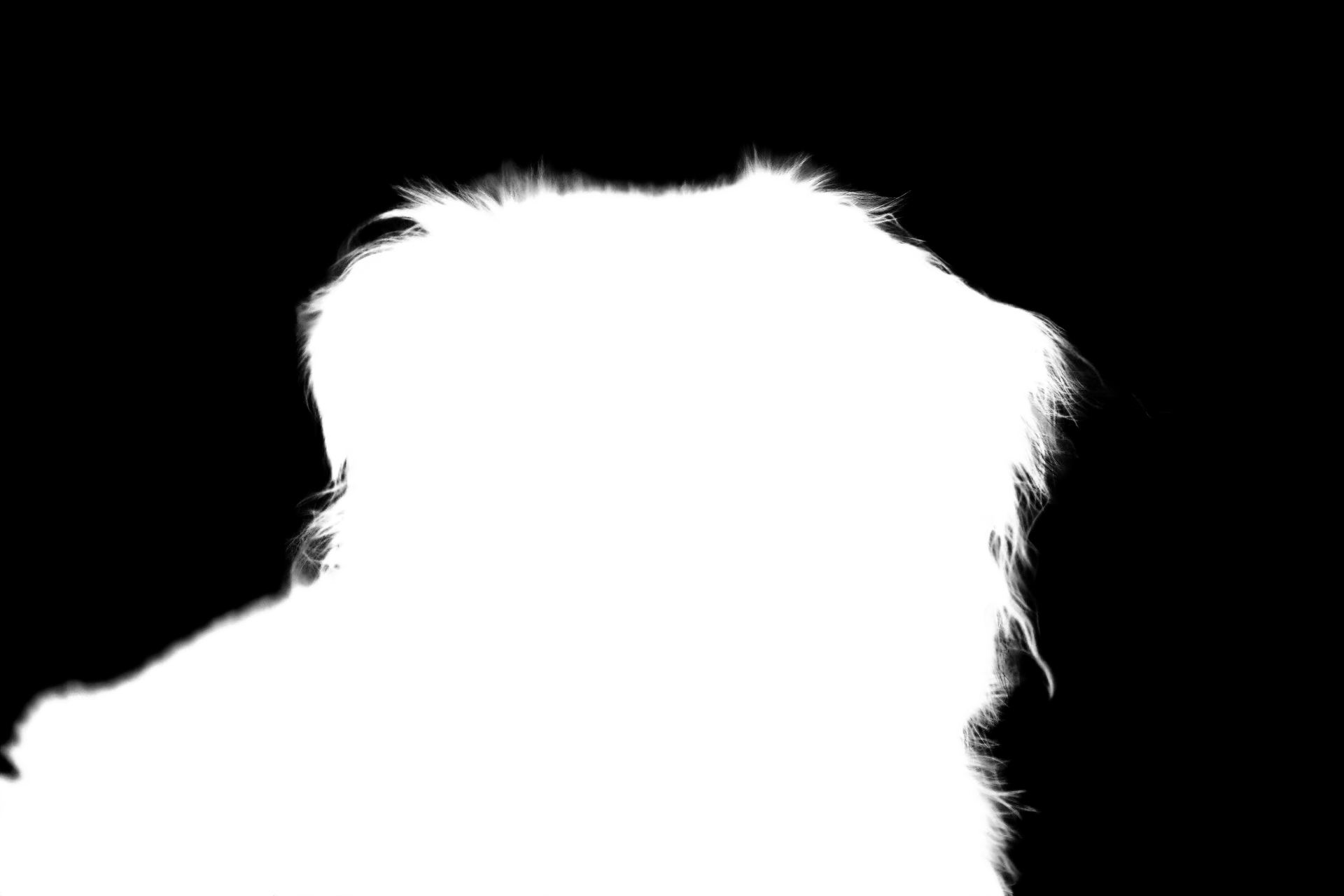} &
				\includegraphics[height=0.108\linewidth]{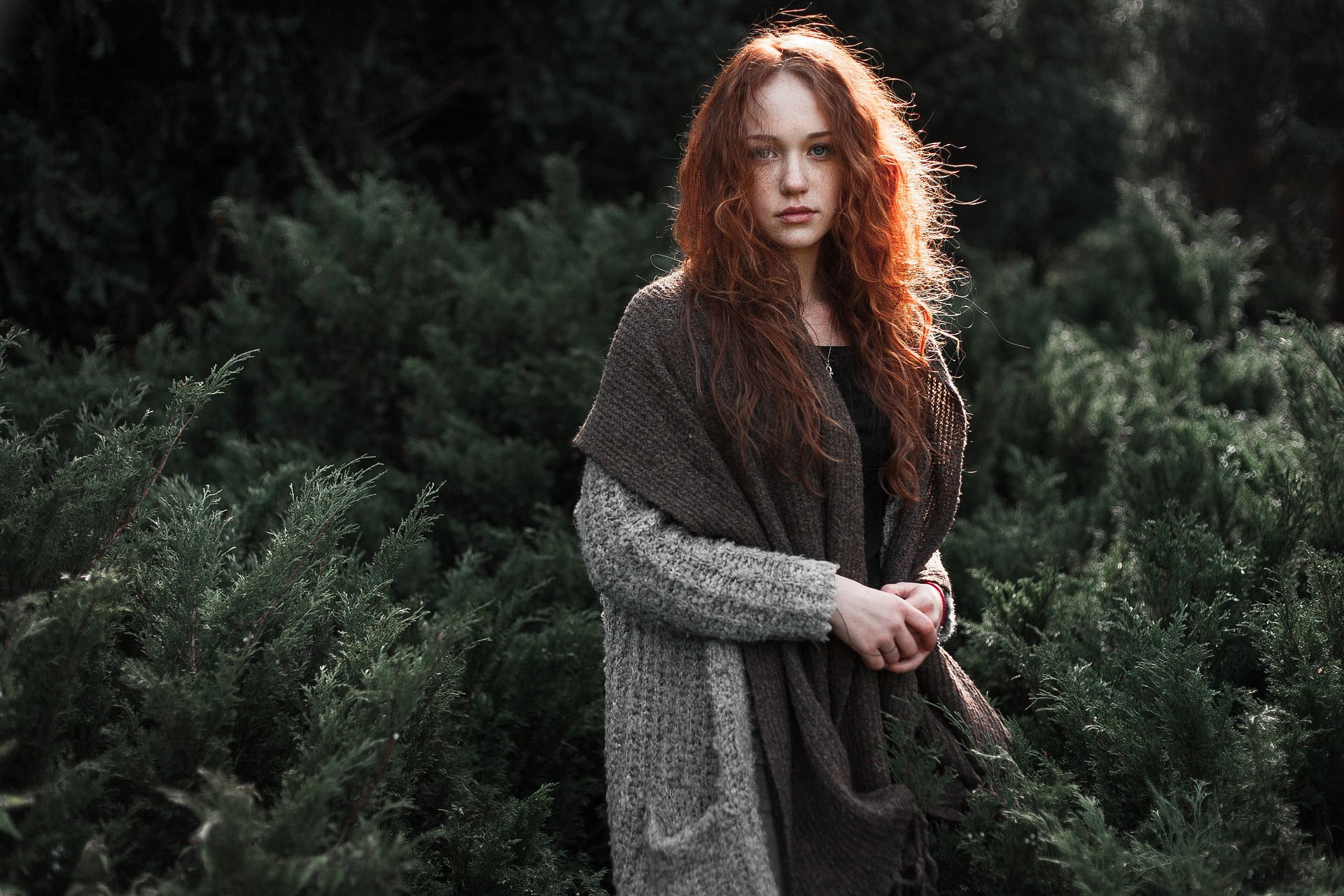} &
				\includegraphics[height=0.108\linewidth]{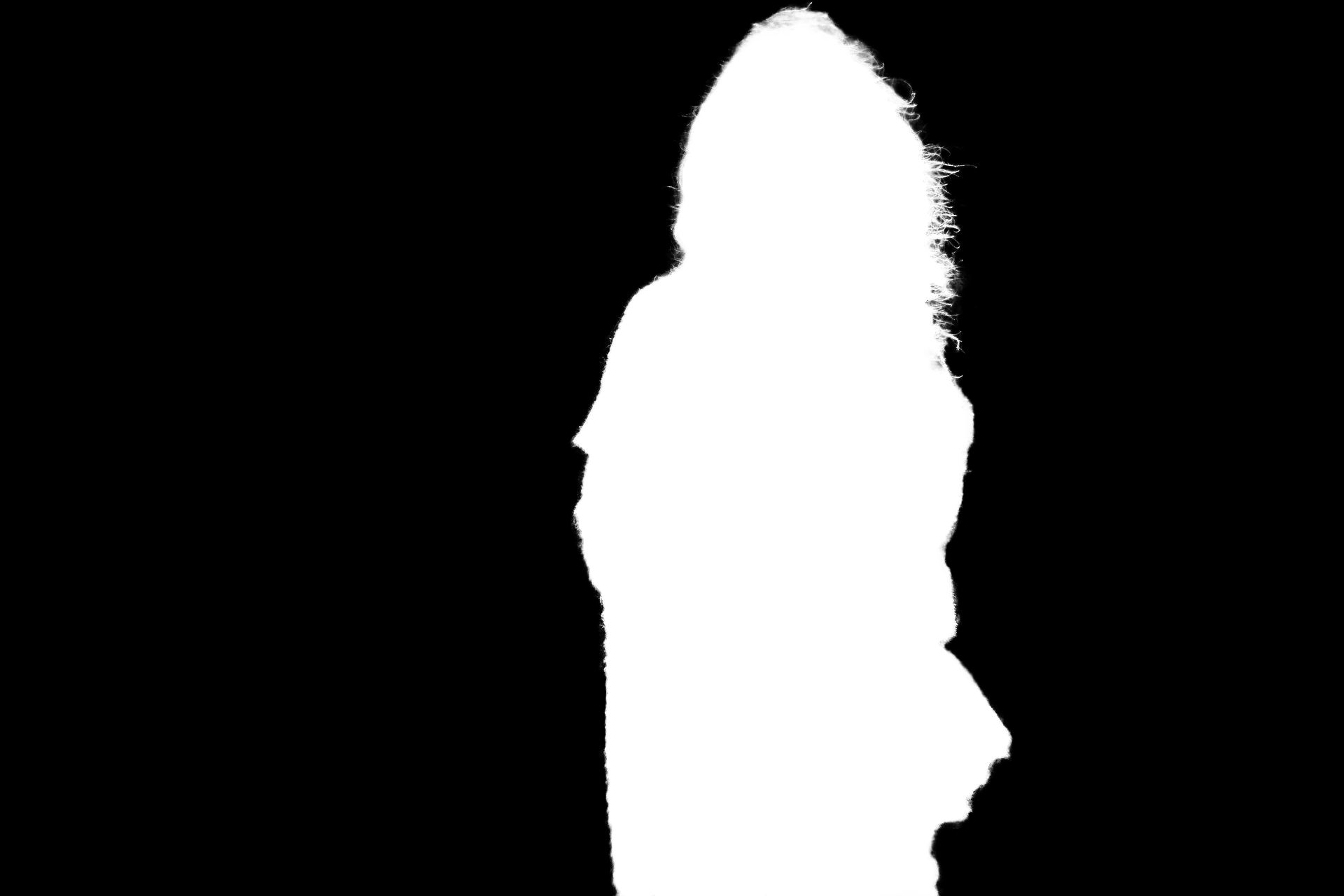} &
				\includegraphics[height=0.108\linewidth]{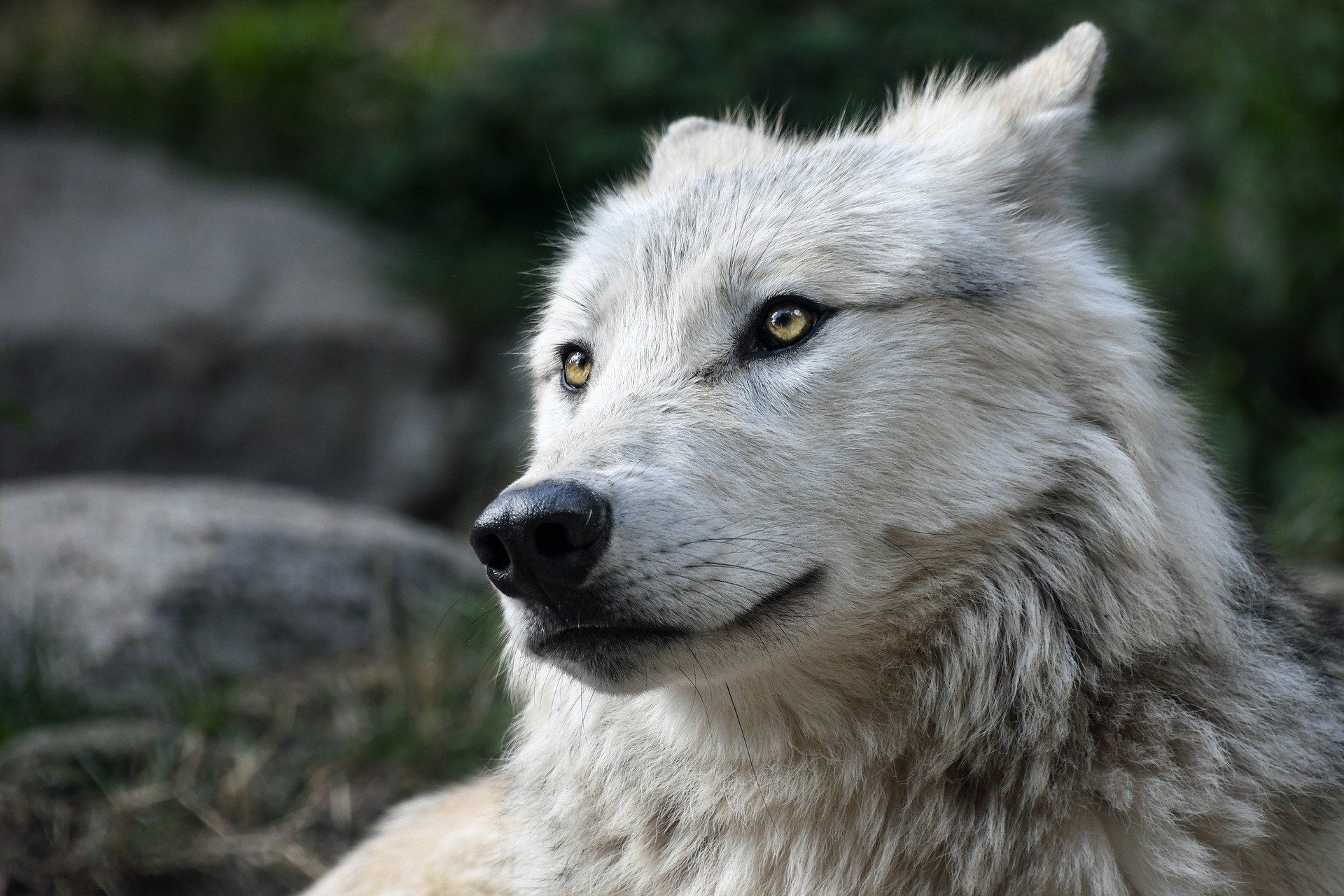} &
				\includegraphics[height=0.108\linewidth]{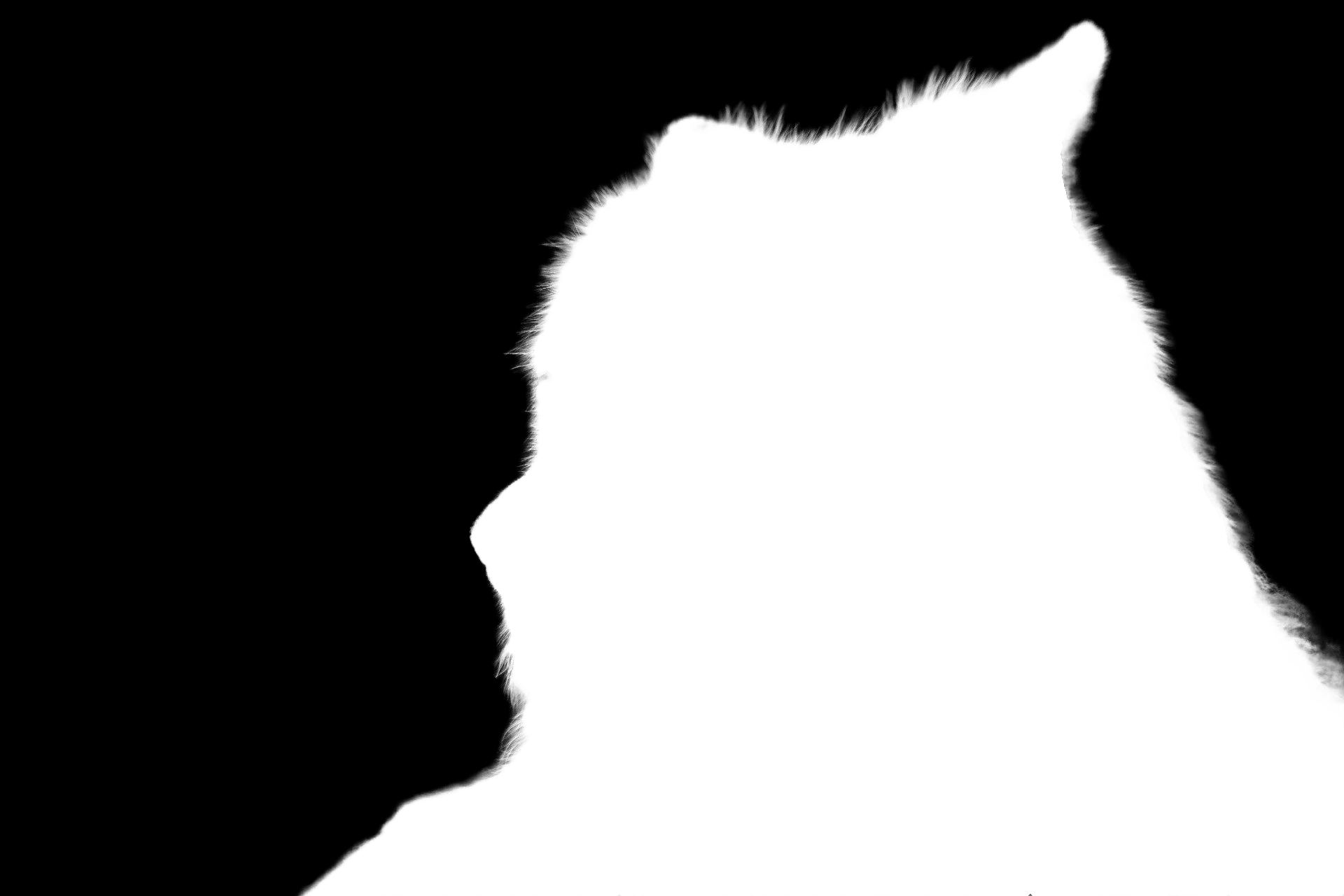} \\
		\end{tabular}}
	\end{center}
	\vspace{-5mm}
	\caption{The alpha mattes generated by I2GFP on the real-world images.}
	\vspace{-3mm}
	\label{fig:natural}
\end{figure}

\vspace{-2mm}
\subsection{Results on the Alpha Matting Benchmark}
\label{ssec:bench}
\vspace{-1mm}
The visual results of the benchmark dataset are shown in Fig.~\ref{fig:alpha}. Without tricks, we can achieve the overall rank of $11.8$ on this list, which ranks seventh among all published papers. Only our model employs a simple backbone to extract features, and the performance promotion is primarily from the well-designed intensive connections and global foreground perception. The alpha features from different levels share profound communications, and necessary foreground details are supplied to refine the boundaries and textures.

\vspace{-2mm}
\subsection{Results on Natural Images}
\label{ssec:natural}
\vspace{-1mm}
Here we display additional results on natural images (Fig.~\ref{fig:natural}). I2GFP can predict high-quality alpha mattes with complex backgrounds (the color disturbance and blur). The transition regions have a clear distinction to separate the foreground objects, which is mostly benefited from our wider and higher feature representation. Notably, the global foreground perception branch can provide essential appearances even affected by illumination (second example in Fig.~\ref{fig:natural}). The trimaps are generated by off-the-shelf segmentation models, and the dependence on high-quality trimap is a problem existing in most current matting methods. However, trimaps can define the user interests, which restricts the valid applications for some trimap-free methods~\cite{Qiao_2020_CVPR,Zhang2019CVPR}.

\vspace{-2mm}
\section{Conclusions and Future Works}
\label{sec:conclusion}
\vspace{-1mm}
In this paper, we review most existing matting architectures and introduce our motivation for wider and higher feature representation. We propose the Intensive Integration and Global Foreground Perception network (I2GFP) to achieve extensive feature representation, perception, and communication for alpha mattes. IC can promote feature communication between upsampling layers and different-level encoder attributes, while GFP branch can capture rich appearances with large kernels. We perform extensive experiments on public datasets and natural images. The results show that I2GFP can achieve state-of-the-art performance, proving the effectiveness of IC, GFP, and the wider and higher feature fields motivation. Besides, we will explore more flexible semantic modules to replace the existing pre-trained backbone in the future, which we believe can achieve a better performance/capacity trade-off.

\vspace{-3mm}

\end{document}